\documentclass[10pt,twocolumn,letterpaper]{article}

\usepackage{paperstyle}

\makeatletter
\@ifundefined{Gread@transgrouptrue}{}{\let\Gread@transgrouptrue\Gread@transgroupfalse}
\makeatother

\definecolor{linkblue}{rgb}{0.21,0.49,0.74}
\usepackage[pagebackref,breaklinks,colorlinks,allcolors=linkblue]{hyperref}

\makeatletter
\newwrite\supp@auxout
\def\@inputsuppaux#1{\@input{#1}}
\newcommand{\SuppRefsStart}{%
  \gdef\@extra@b@citeb{@supp}%
  \if@filesw
    \immediate\openout\supp@auxout=\jobname-supp.aux\relax
    \immediate\write\supp@auxout{\relax}%
    \immediate\write\supp@auxout{\string\gdef\string\@extra@binfo{@supp}}%
    \immediate\write\@mainaux{\string\@inputsuppaux{\jobname-supp.aux}}%
    \global\let\@auxout\supp@auxout
  \fi}
\newcommand{\SuppReferences}[2]{%
  \if@filesw
    \immediate\write\supp@auxout{\string\bibstyle{#1}}%
    \immediate\write\supp@auxout{\string\bibdata{#2}}%
  \fi
  \@input@{\jobname-supp.bbl}%
  \if@filesw\immediate\write\supp@auxout{\string\gdef\string\@extra@binfo{}}\fi}
\AddToHook{enddocument/afterlastpage}{%
  \if@filesw
    \global\let\@auxout\@mainaux
    \immediate\closeout\supp@auxout
  \fi}
\makeatother

\AtBeginDocument{\widowpenalty=10000 \clubpenalty=10000 \displaywidowpenalty=10000 }

\title{HuluOPD: Unifying Heterogeneous Medical Specialists\\into One Lightweight Model with Unlabeled Prompts}

\author{%
Yuncheng Zhang$^{1}$,
Songtao Jiang$^{1}$,
Yitian Tang$^{1}$,
Zijie Meng$^{1}$,
Zihan Xu$^{1}$,\\
Xuxiang Ren$^{2}$,
Jian Wu$^{1}$,
Hualiang Wang$^{1,*}$,
Zuozhu Liu$^{1,*}$\\[0.6em]
$^{1}$Zhejiang University\qquad
$^{2}$University of Hong Kong
}

\begin{document}
\maketitle
\AddToHookNext{shipout/foreground}{%
  \put(\LenToUnit{\dimexpr 1in+\oddsidemargin\relax},\LenToUnit{\dimexpr -1in-\topmargin-\headheight-\headsep-\textheight-10pt\relax}){%
    \makebox[0pt][l]{\rule{0.4\columnwidth}{0.4pt}}}%
  \put(\LenToUnit{\dimexpr 1in+\oddsidemargin+1em\relax},\LenToUnit{\dimexpr -1in-\topmargin-\headheight-\headsep-\textheight-19pt\relax}){%
    \makebox[0pt][l]{\footnotesize$^{*}$Corresponding authors.}}}
\begin{abstract}
Specialist medical vision-language models (VLMs) are strong within their specialties but hard to unify: their training data are often private, and independently developed specialists cannot be assumed to share an architecture. Deploying these large models separately is costly. We present \textbf{HuluOPD}, an on-policy distillation recipe that unifies heterogeneous specialists into one lightweight VLM using only unlabeled prompts, without requiring reference answers or the specialists' training data. The lightweight model answers each prompt, and the specialist of the prompt's modality supervises its response; a continuation-aligned cross-tokenizer loss makes this possible even when they differ in architecture and tokenizer. We apply HuluOPD to five specialists, for ultrasound, pathology, optical coherence tomography, fundus photography, and dental imaging. To distill and evaluate them, we build HuluOPDCorpus, 914k unlabeled prompts, and HuluOPDBench, 47k held-out test items, from over 90 public datasets, and score all models under one protocol on 132k test items, including third-party benchmarks. A single model with 10\% of the five specialists' combined parameters retains 94\% of their question-answering performance, and one with 2\% retains 89\%. To our knowledge, this is the first demonstration of unifying heterogeneous medical specialist VLMs without requiring their training data or a shared architecture.

\end{abstract}
\section{Introduction}
\label{sec:intro}

\begin{figure}[t]
  \centering
  \includegraphics[width=\linewidth]{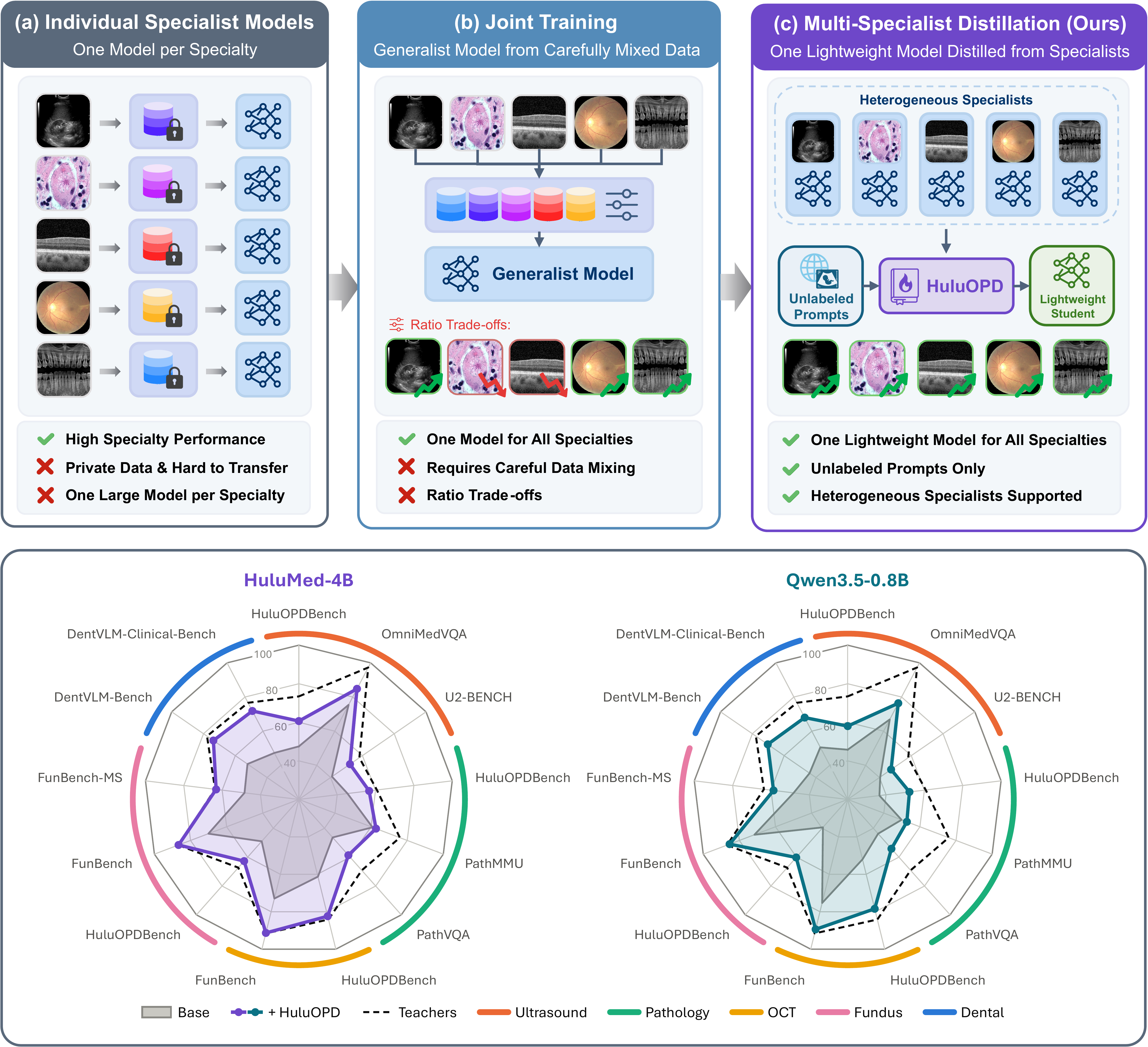}
  \caption{\textbf{Three ways to serve many specialties, and our results.}
  \textbf{(a)}~One large specialist per specialty, trained on private data.
  \textbf{(b)}~One generalist trained on pooled labeled data (arrows are illustrative).
  \textbf{(c)}~HuluOPD distills heterogeneous specialists into one lightweight student with unlabeled prompts only.
  \textbf{Bottom:} the two unified students on 13 VQA benchmarks (\cref{tab:unified}).}
  \label{fig:teaser}
\end{figure}

\begin{figure*}[t]
  \centering
  \includegraphics[width=\textwidth]{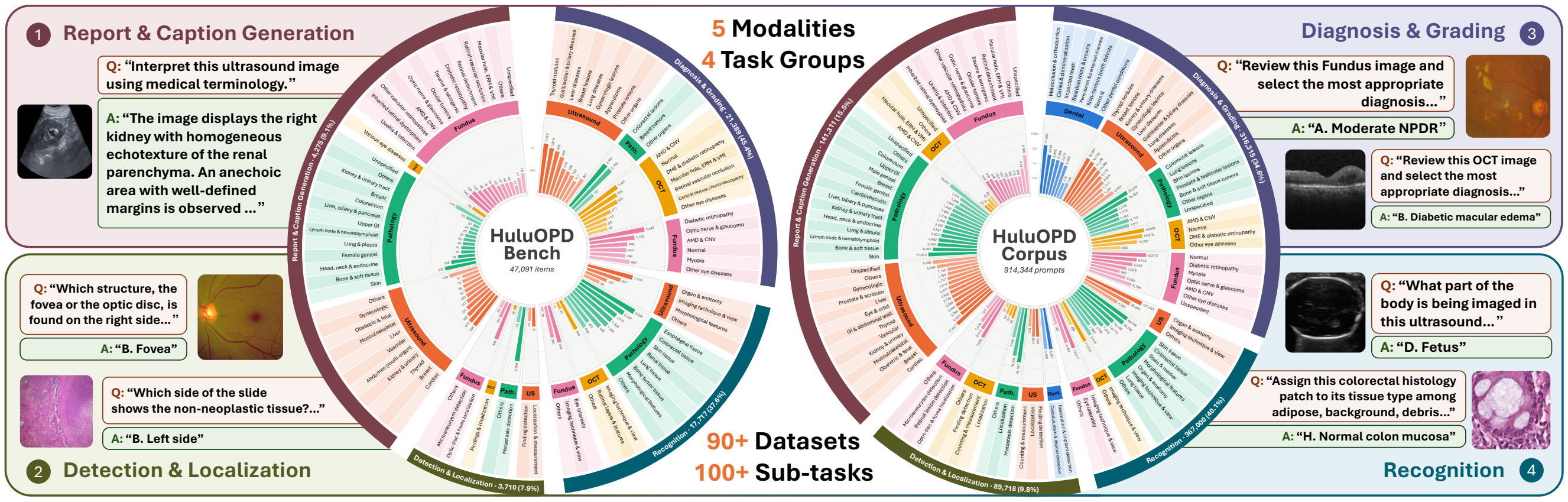}
  \caption{\textbf{HuluOPDBench and HuluOPDCorpus.} Outer arcs: the four task groups; colored ring: imaging modality; bars: number of items per sub-task (log scale). Sources: Appendix~\cref{tab:supp_src_corpus,tab:supp_src_bench}; dental imaging is evaluated on the DentVLM benchmarks.}
  \label{fig:data}
\end{figure*}

A common way to build medical vision-language models (VLMs) is to train one model per specialty, such as ultrasound~\cite{she2026echovlm,guo2024llavaultra}, pathology~\cite{lu2024pathchat,sun2025cpathomni}, ophthalmology~\cite{deng2026fundusr1,wu2026ophthareason}, or dentistry~\cite{meng2025dentvlm,hao2025oralgptomni} (\cref{fig:teaser}a).
Each such model performs strongly within its specialty but covers only that specialty's modality and question types, so serving many specialties requires many large models.
Their labeled training data, curated with substantial clinician effort, are rarely released for ethical and privacy reasons, leaving the weights as the main carrier of this expertise.
Moreover, specialist teams develop their models independently, without coordinating on a shared architecture or tokenizer.

Recent work instead trains one generalist on labeled data pooled across imaging modalities~\cite{lin2025healthgpt,sellergren2025medgemma,xu2025lingshu,jiang2025hulumed} (\cref{fig:teaser}b).
This route requires labeled data for every target specialty, yet such data are rarely released.
Such a model takes all of its knowledge from this pool, so what it learns about each specialty is decided by how the pool is composed.
Because medical specialties differ widely in images, knowledge, and questions, the mixture has to be balanced carefully, and a larger share for one specialty leaves a smaller one for another.
In practice, generalists still tend to trail the specialists within their own \mbox{specialties}~\cite{meng2025dentvlm,sun2025cpathomni,wu2026ophthareason}.

Unifying specialist expertise should require neither access to private training data nor prior coordination on model architecture.
We pursue this goal by distilling multiple specialists into one lightweight student VLM using only unlabeled prompts, without ground-truth answers (\cref{fig:teaser}c).
This setting remains underexplored on two fronts: data resources and distillation algorithms.

\textbf{On the data side,} specialty-specific datasets provide no common basis for distilling heterogeneous specialists and measuring knowledge transfer across modalities and tasks.
We therefore organize more than 90 public datasets into \textbf{HuluOPDCorpus}, 914k unlabeled image--question prompts, and \textbf{HuluOPDBench}, 47k held-out test items (\cref{fig:data}), spanning four task groups and more than 100 fine-grained sub-tasks.
Reference answers are never used for distillation, and image-level duplicate and near-duplicate checks separate the corpus prompts from benchmark test images.
With seven third-party benchmarks, evaluation covers five imaging modalities and 132k test items under one protocol.

\textbf{On the algorithmic side,} distillation must accommodate each specialist's architecture and tokenizer; differing vocabularies prevent token-level comparison with the student.
Off-policy distillation uses pre-generated teacher responses~\cite{kim2016sequence}, but creates exposure bias: training uses teacher prefixes whereas inference uses the student's own~\cite{bengio2015scheduled,agarwal2024gkd}.
On-policy distillation (OPD)~\cite{agarwal2024gkd,gu2024minillm,lu2025onpolicy} instead supervises responses generated by the student.
However, most OPD methods, including Med\mbox{-}OPD~\cite{qian2026medopd}, assume a shared tokenizer~\cite{yang2025qwen3,xiaomi2026mimo,ma2026mopd}; existing cross-tokenizer variants~\cite{gold2025,sun2026simct,wang2026bpm} target text-only, single-teacher settings.
Whether OPD can unify multiple heterogeneous medical specialist VLMs therefore remains open.

\begin{figure*}[t]
  \centering
  \includegraphics[width=\linewidth]{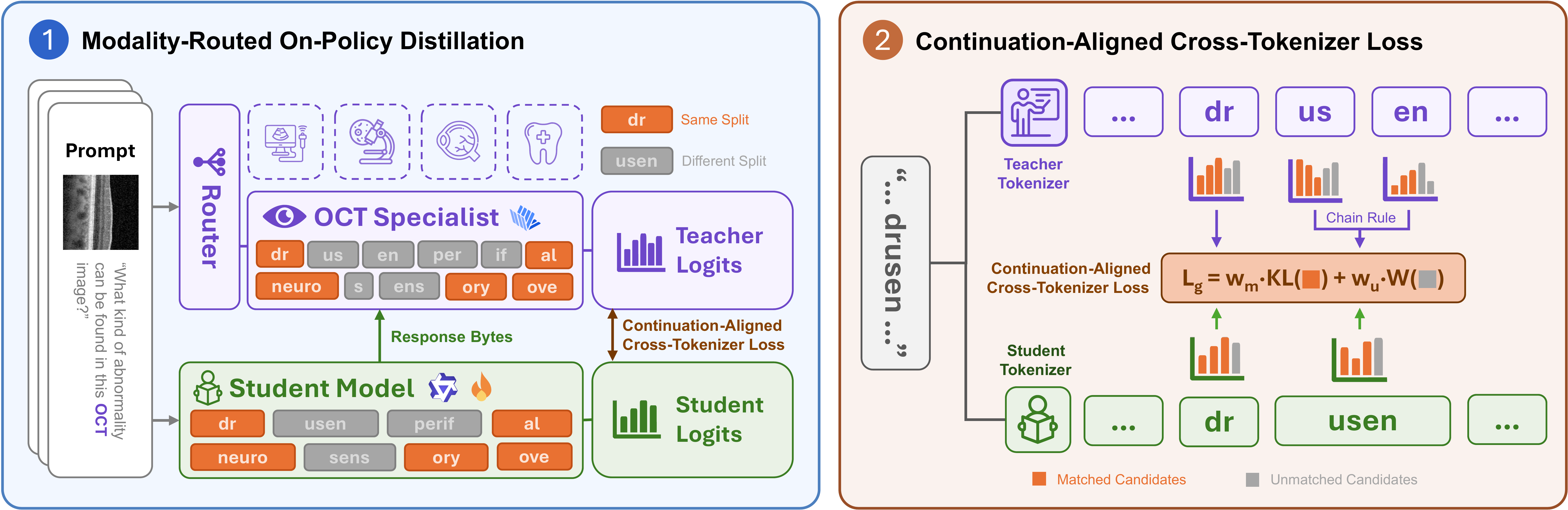}
  \caption{\textbf{Overview of HuluOPD.}
  \textbf{(1)}~A router sends each prompt to the specialist of its modality (here, OCT), which re-tokenizes the student's response; orange tokens are split alike by both tokenizers, gray ones differently.
  \textbf{(2)}~Candidates are paired by the text they complete (\eg, teacher \emph{us}+\emph{en} with student \emph{usen}); matched ones (orange) enter KL, unmatched ones (gray) a sorted-probability term $W$.}
  \label{fig:method}
\end{figure*}

We present \textbf{HuluOPD}, a multi-teacher OPD recipe that distills heterogeneous medical specialists into one lightweight student using only unlabeled prompts, without requiring the specialists' training data.
After a brief warm-up on cached teacher responses, the student answers each prompt, and the frozen specialist teacher of the prompt's modality reads the response with its own processor and tokenizer and returns its next-token distributions as supervision (\cref{fig:method}).
Since two tokenizers can split a clinical term into different subwords, the same token can complete different text in the two models; our continuation-aligned cross-tokenizer loss therefore pairs teacher and student candidates by the text they complete.

We distill five specialists from three VLM families, for ultrasound, pathology, optical coherence tomography (OCT), fundus photography, and dental imaging, into two students, HuluMed\mbox{-}4B~\cite{jiang2025hulumed} and Qwen3.5\mbox{-}0.8B~\cite{qwen35}, about $2\times$ and $10\times$ smaller than the teachers.
Distilled from one specialist at a time, the two students rise from 75\% and 69\% to 93\% and 90\% of the teachers' question-answering score, respectively (\cref{tab:separate}).
Distilled from all five at once, a single student reaches 94\% and 89\% of the teachers' score with only 10\% and 2\% of the five specialists' combined parameters (\cref{fig:teaser}, bottom; \cref{tab:unified}).
To our knowledge, this is the first evidence that heterogeneous medical specialist VLMs can be unified into one lightweight model without requiring their training data or a shared architecture.

\noindent Our contributions are summarized as follows:
\begin{itemize}
  \item \textbf{Resources for multi-specialist distillation.} We organize more than 90 public datasets into HuluOPDCorpus, unlabeled distillation prompts, and HuluOPDBench, held-out tests; with seven third-party benchmarks, they cover five modalities and four task groups (\cref{fig:data}).
  \item \textbf{Heterogeneous multi-specialist OPD.} HuluOPD unifies multiple specialists into one lightweight VLM from unlabeled prompts alone. Each specialist keeps its own processor and tokenizer, and a continuation-aligned cross-tokenizer loss bridges the vocabulary gap, so the recipe applies to independently developed specialists.
  \item \textbf{Strong lightweight students.} One student reaches 94\% of the specialists' question-answering performance with 10\% of their combined parameters, and 89\% with 2\%.
\end{itemize}

\section{Related Work}
\label{sec:related}

\paragraph{Medical VLMs.}
Specialist VLMs exist for many specialties, such as ultrasound~\cite{guo2024llavaultra,she2026echovlm}, pathology~\cite{lu2024pathchat,sun2025cpathomni}, ophthalmology~\cite{wu2026ophthareason,deng2026fundusr1}, and dentistry~\cite{meng2025dentvlm,hao2025oralgptomni}, while generalist medical VLMs learn from data pooled across specialties~\cite{li2023llavamed,sellergren2025medgemma,lin2025healthgpt,jiang2025hulumed}.
Existing specialists can also be combined without retraining, either by invoking them as tools at inference ~\cite{nath2025vilam3}, or by merging them in parameter space~\cite{mou2026mergemedbench}. The former keeps every specialist deployed, while the latter requires them to share a base model. HuluOPD instead distills them into one lightweight VLM, even when they share no base model.

\paragraph{On-policy distillation.}
OPD~\cite{agarwal2024gkd,gu2024minillm,lu2025onpolicy} is now commonly used to distill large models into lightweight ones~\cite{yang2025qwen3} and has been extended to VLMs~\cite{bousselham2026vold}, including medical ones~\cite{qian2026medopd}.
With several teachers, most methods route each prompt to the teacher of its domain~\cite{xiaomi2026mimo,ma2026mopd,zhao2026opod,qin2026videomopd}; these teachers are often copies of the student fine-tuned per domain, and several methods also use reference answers or rewards.
HuluOPD also routes each prompt to one teacher, but its teachers are separately trained specialists, and it uses neither answers nor rewards.

\paragraph{Cross-tokenizer distillation.}
OPD matches the student's next-token distribution to the teacher's at every position, which assumes that the two models share a vocabulary.
Cross-tokenizer methods remove this assumption.
Off-policy ones compare sorted probabilities~\cite{boizard2024uld}, map or project tokens between vocabularies~\cite{wan2024fusellm,sreenivas2026xtoken}, or match aligned text chunks~\cite{minixhofer2025alm}; on-policy ones pair identical token strings~\cite{gold2025}, compare text spans that both tokenizers can produce~\cite{niu2026breaking,sun2026simct}, or marginalize over byte prefixes~\cite{wang2026bpm}.
HuluOPD pairs a teacher candidate with a student candidate only when the two complete the same text (\cref{sec:loss}); \cref{tab:capability} compares it with these methods.
\section{Method}
\label{sec:method}

HuluOPD combines three components (\cref{fig:method}): an off-policy warm-up on cached teacher responses (\cref{sec:warmup}), on-policy distillation that routes each prompt to the teacher of its modality (\cref{sec:routing}), and a cross-tokenizer loss that pairs the candidates of the two models by the text they complete (\cref{sec:loss}).

\subsection{Problem Formulation}
\label{sec:formulation}

Let $k\in\{1,\dots,K\}$ index the imaging modalities.
For each modality, we are given a frozen specialist teacher $T_k$ and a pool $\mathcal{D}_k$ of unlabeled prompts, each consisting of an image and a question with no reference answer.
We train one student $S_\theta$ to answer every prompt $x\in\mathcal{D}=\bigcup_k\mathcal{D}_k$ as the teacher of its modality, $T_{k(x)}$, would; the modality $k(x)$ is known from the data source of $x$.
We call the case $K=1$ single-specialist distillation and the case $K>1$ unified multi-specialist distillation.

When teacher and student share a tokenizer, OPD~\cite{agarwal2024gkd,lu2025onpolicy} samples a response $y=(y_1,\dots,y_L)\sim S_\theta(\cdot\mid x)$ and matches the next-token distributions of the two models on the student's own prefixes:
\begin{equation}
  \mathcal{L}_{\mathrm{OPD}}(\theta)=
  \mathop{\mathbb{E}}_{\substack{x\sim\mathcal{D}\\ y\sim S_\theta(\cdot\mid x)}}
  \Big[\frac{1}{L}\sum_{n=1}^{L}
  \mathrm{KL}\big(p_T^{\,n}\,\big\|\,p_S^{\,n}\big)\Big],
  \label{eq:opd}
\end{equation}
where $p_T^{\,n}=p_T(\cdot\mid x,y_{<n})$ and $p_S^{\,n}=p_S(\cdot\mid x,y_{<n})$ are the teacher's and student's next-token distributions after the prefix $y_{<n}$.
We use the forward KL and treat the sampled response as fixed, so gradients flow only through $p_S$~\cite{agarwal2024gkd}.
Because the supervision is the teacher's distribution rather than a reference answer, only prompts are needed.

In our setting, teachers and student may differ in vision encoder, language backbone, and tokenizer.
The first two leave \cref{eq:opd} intact, as each model reads the image and question with its own processor and chat template.
A different tokenizer does not: when the teacher vocabulary $\mathcal{V}_k$ differs from the student vocabulary $\mathcal{V}_S$, the two models can split a response differently, and their distributions no longer correspond entry by entry.
We therefore connect them through the bytes of the response, which both tokenizers encode losslessly; $\operatorname{txt}(v)$ denotes the bytes of token $v$ and $\oplus$ concatenation.

\subsection{Off-Policy Warm-Up}
\label{sec:warmup}

A base student answers in its own format, length, and style, and on such responses much of the teacher's signal would go to correcting the format rather than the content.
Following the warm-up-then-OPD schedule of recent model families~\cite{yang2025qwen3,bai2025qwen3vl}, we therefore first fine-tune the student on teacher responses: for each modality, the teacher answers a stratified subset $\mathcal{W}_k\subset\mathcal{D}_k$ of the prompts once (Appendix~\ref{sec:supp_data}), and the student is fine-tuned on these cached responses, encoded with its own tokenizer:
\begin{equation}
  \mathcal{L}_{\mathrm{warm}}(\theta)=
  -\mathop{\mathbb{E}}_{x\sim\mathcal{W}}\sum_{n=1}^{|\hat y|}\log p_S(\hat y_n\mid x,\hat y_{<n}),
  \label{eq:warmup}
\end{equation}
where $\hat y$ is the cached response of teacher $T_{k(x)}$; as it reaches the student as text, this stage needs no vocabulary alignment.
In unified multi-specialist distillation, $\mathcal{W}$ mixes the cached responses of all $K$ teachers, so that one student learns each teacher's answer format on its modality.

\subsection{Modality-Routed On-Policy Distillation}
\label{sec:routing}

After the warm-up, training is purely on-policy (\cref{fig:method}, left): for a prompt $x$, the current student samples a response $y$ (decoding settings in Appendix~\cref{tab:supp_hparams}), and no teacher response or reference answer enters the loss.
We keep the sampled tokens unchanged and write $b$ for their bytes.

A router sends $x$ and $y$ to the teacher of the prompt's modality, $T=T_{k(x)}$, which renders the image and question with its own processor and chat template, encodes $b$ with its own tokenizer into $z=(z_1,\dots,z_{L'})$, and returns $p_T^{\,m}=p_T(\cdot\mid x,z_{<m})$ at every position $m$ in one teacher-forced forward pass; it never generates.

The student minimizes the batch mean of the per-sample loss $\mathcal{L}(x,y)$ of \cref{sec:loss}, so all responses weigh the same regardless of length.

In unified multi-specialist distillation, every batch mixes prompts of all $K$ modalities, each scored by its own teacher, so every optimizer step draws on all specialists.

\subsection{Continuation-Aligned Cross-Tokenizer Loss}
\label{sec:loss}

The loss compares the student response $y$ with the teacher's encoding $z$ of the same bytes $b$ in three steps (\cref{fig:method}, right): it aligns and merges, pairs, and compares; we drop the modality index and write $\mathcal{V}_T$ for the teacher vocabulary.
Our running example is \emph{drusen}, which the student splits into \emph{dr}+\emph{usen} and the teacher into \emph{dr}+\emph{us}+\emph{en}.

\paragraph{Alignment and merging.}
We align the two tokenizations by bytes and merge each side's predictions within the aligned spans~\cite{minixhofer2025alm,gold2025}.
Cutting $b$ wherever both tokenizations have a token boundary splits the response into $G$ groups; group $g$ holds the student tokens with indices $\mathcal{I}_g$ and the teacher tokens with indices $\mathcal{J}_g$, which cover the same bytes, such as \emph{usen} and \emph{us}+\emph{en}.
If the student ended the response itself, its stop token and the teacher's end-of-sequence token form a final group.

Within a group, each side's predictions are merged by the chain rule in the log domain.
Let $n_g$ be the position of the last student token of group $g$ and $\delta^S_g$ the bytes of the student's earlier tokens in the group.
For every candidate $v\in\mathcal{V}_S$,
\begin{equation}
  \log P^S_g(v)=\sum_{n\in\mathcal{I}_g,\,n<n_g}\log p_S^{\,n}(y_n)\;+\;\log p_S^{\,n_g}(v)
  \label{eq:merge}
\end{equation}
is the log-probability that the student, from the start of the group, emits these earlier tokens and then $v$, a token path that continues the response with the text $\delta^S_g\oplus\operatorname{txt}(v)$.
The teacher's $\log P^T_g(u)$, $u\in\mathcal{V}_T$, and $\delta^T_g$ are defined in the same way on its encoding $z$.

\paragraph{Pairing by continuation.}
Two candidates are comparable only when they continue the response with the same text, so we pair a student candidate $v\in\mathcal{V}_S$ with a teacher candidate $u\in\mathcal{V}_T$ exactly when their continuations are identical:
\begin{equation}
  \mathcal{M}_g=\big\{(v,u):\ \delta^S_g\oplus\operatorname{txt}(v)=\delta^T_g\oplus\operatorname{txt}(u)\big\}.
  \label{eq:pairing}
\end{equation}
For \emph{usen} and \emph{us}+\emph{en}, $\delta^S_g$ is empty and $\delta^T_g$ is \emph{us}, so the teacher's candidate \emph{en} stands for \emph{usen} and is paired with the student's \emph{usen}.
We call the paired candidates matched and write $\mathcal{U}^S_g$ and $\mathcal{U}^T_g$ for the unmatched ones, which include any candidate whose text is only a prefix or an extension of one on the other side.

\paragraph{Hybrid loss.}
The loss of group $g$ combines two terms \cite{gold2025}: a forward-KL term over the matched pairs (orange in \cref{fig:method}, right) and, for the unmatched candidates (gray), which have no counterpart, the distance $W$ between sorted probabilities of ULD~\cite{boizard2024uld}:
\begin{equation}
\begin{aligned}
  \mathcal{L}_g={}& w_m\!\!\!\sum_{(v,u)\in\mathcal{M}_g}\!\!\! P^T_g(u)\big[\log P^T_g(u)-\log P^S_g(v)\big]\\
  &+w_u\,W\big(P^S_g[\mathcal{U}^S_g],\,P^T_g[\mathcal{U}^T_g]\big),
\end{aligned}
\label{eq:hybrid}
\end{equation}
where $W(p,q)=\|\operatorname{sort}(p)-\operatorname{sort}(q)\|_1$ sorts each vector in decreasing order and zero-pads the shorter one, so it needs no correspondence between candidates.
The weights are fixed by the global vocabulary overlap: $w_m=|\mathcal{M}|/|\mathcal{V}_T|$ and $w_u=1-w_m$, where $\mathcal{M}$ is the set of token pairs with identical strings in the two vocabularies.
When teacher and student share a tokenizer, the teacher's encoding $z$ usually reproduces the student's tokens $y$: each group then holds one token per side, all candidates are matched, $w_m\approx1$, and \cref{eq:hybrid} reduces to the per-token forward KL of \cref{eq:opd}.
The same loss therefore serves teachers with and without the student's tokenizer.
The per-sample loss averages over groups, $\mathcal{L}(x,y)=\frac{1}{G}\sum_{g=1}^{G}\mathcal{L}_g$, and is computed in the log domain without floors or clamps (Appendix~\ref{sec:supp_impl}).

\section{Experiments}
\label{sec:experiments}

\subsection{Datasets and Evaluation}
\label{sec:data}

\paragraph{HuluOPDCorpus and HuluOPDBench.}
We build two datasets for heterogeneous specialist distillation (\cref{fig:data}).
HuluOPDCorpus holds 914k unlabeled image--question prompts for the five modalities, whose reference answers are never used in training, and HuluOPDBench holds 47k test items for ultrasound, pathology, OCT, and fundus photography from held-out test splits.
Both draw on more than 90 public datasets, and span four task groups: diagnosis and grading, recognition, detection and localization, and report and caption generation (Appendix~\cref{sec:supp_datasets,tab:supp_comp,tab:supp_src_corpus,tab:supp_src_bench}).
Both are organized into more than 100 fine-grained sub-tasks, including many small and niche ones (\cref{fig:data}); this broad coverage makes HuluOPDBench a challenging benchmark.
We verified image by image that no image leaks from HuluOPDCorpus into HuluOPDBench: none of its test images has a duplicate or near-duplicate among the prompts, so its scores are measured on images outside the distillation data.
HuluOPDBench has no dental part: dental imaging is evaluated on the DentVLM benchmarks~\cite{meng2025dentvlm}, which we count among the third-party ones.

\newlength{\gw}\settowidth{\gw}{\tiny +00.00}
\newcommand{\g}[2]{#1\,\makebox[\gw][l]{\tiny\textcolor{ForestGreen}{#2}}}
\newcommand{\gna}{\ooalign{\hphantom{00.00}\cr\hfil--\hfil\cr}\,\hphantom{\makebox[\gw]{}}}
\newcommand{\modrow}[1]{\multicolumn{7}{@{}l@{}}{\makebox[0pt][l]{\setlength{\fboxsep}{0pt}\colorbox{black!7}{\makebox[\linewidth][l]{\rule[-2.2pt]{0pt}{8.2pt}\hspace{1pt}\textit{#1}}}}}}
\newcommand{\modrowN}[2]{\multicolumn{#1}{@{}l@{}}{\makebox[0pt][l]{\setlength{\fboxsep}{0pt}\colorbox{black!7}{\makebox[\linewidth][l]{\rule[-2.2pt]{0pt}{8.2pt}\hspace{1pt}\textit{#2}}}}}}
\newlength{\genw}\newlength{\genwtmp}
\newcommand{\genwmax}[1]{\settowidth{\genwtmp}{#1}\ifdim\genwtmp>\genw\setlength{\genw}{\genwtmp}\fi}
\newcommand{\gn}[1]{\makebox[\genw][c]{#1}}
\newcommand{\gh}[2]{\makebox[\genw][c]{\smash{\shortstack{#1\strut\\[-2.5pt]#2\strut}}}}
\newlength{\genruleraise}\setlength{\genruleraise}{3.9pt}
\newcommand{\genrule}{\leaders\hrule height \dimexpr\genruleraise+\cmidrulewidth\relax depth -\genruleraise\hfill\null}

\paragraph{Third-party benchmarks.}
Besides HuluOPDBench, we evaluate on subsets of OmniMedVQA\mbox{-}US~\cite{hu2024omnimedvqa} and U2\mbox{-}BENCH~\cite{le2026u2bench} for ultrasound, PathMMU~\cite{sun2024pathmmu} and the PathVQA~\cite{he2020pathvqa} test split for pathology, and FunBench~\cite{wei2025funbench} for OCT and fundus photography.
The parts we leave out are mostly tasks that HuluOPD does not target (Appendix~\cref{tab:supp_bench} gives the composition of each subset).
Dental imaging is covered by DentVLM\mbox{-}Bench and DentVLM\mbox{-}Clinical\mbox{-}Bench~\cite{meng2025dentvlm}, with dental radiographs and intraoral photographs.
Together with HuluOPDBench, we evaluate on 131,922 test items.

\paragraph{Metrics.}
Single-choice and open-ended questions (VQA) are scored by accuracy: an LLM judge, Qwen3.8\mbox{-}27B~\cite{qwen38}, decides whether the final answer of a response matches the reference.
For multiple-select questions (VQA\mbox{-}MS), the judge extracts the selected options, and we report the mean F1 between the predicted and reference option sets.
Report and caption generation (Gen) is scored by BERTScore-F1~\cite{zhang2020bertscore} with Longformer\mbox{-}large, and all models answer with greedy decoding (Appendix~\ref{sec:supp_eval}).

\subsection{Models and Implementation}
\label{sec:setup}

Each modality has one specialist teacher (Appendix~\cref{tab:models}).
For dental imaging, it is the released DentVLM~\cite{meng2025dentvlm}.
For the other four modalities, we train our own teachers, because released specialists are mostly built for their own tasks and answer formats, and rarely cover all four task groups in HuluOPDBench.
Each of them is obtained by fully fine-tuning a recent general VLM, Qwen3.5\mbox{-}9B~\cite{qwen35} for ultrasound and pathology and InternVL3.5\mbox{-}8B~\cite{wang2025internvl35} for OCT and fundus photography, for three epochs on 29k--380k curated image--question--answer examples of its modality (Appendix~\ref{sec:supp_impl}).
The resulting teachers are strong specialists: they outperform the generalist medical VLMs of similar or larger size that we evaluate on every benchmark, and they are far ahead of both students before distillation (\cref{tab:unified}).

The students are HuluMed\mbox{-}4B, the medical generalist above, and Qwen3.5\mbox{-}0.8B~\cite{qwen35}, a general-purpose VLM, about $2\times$ and $10\times$ smaller than the teachers.
HuluMed\mbox{-}4B shares its architecture with none of the teachers and is distilled across tokenizers from the ultrasound and pathology teachers.
Qwen3.5\mbox{-}0.8B is distilled across tokenizers from the OCT, fundus, and dental teachers; the ultrasound and pathology teachers are larger models of its own family (Appendix~\cref{tab:models}).

For the warm-up (\cref{sec:warmup}), each teacher answers 10k--50k prompts of its modality (Appendix~\cref{tab:supp_budget}), and the student is fine-tuned on these responses for one epoch.
On-policy distillation (\cref{sec:routing}) then runs for two epochs: over all HuluOPDCorpus prompts of the modality in the single-specialist setting (\cref{sec:exp_one_to_one}), and over a 344k-prompt subset across the five modalities in the unified setting (\cref{sec:exp_five_to_one}).
Appendix~\cref{tab:supp_hparams} lists the settings of both stages.

\subsection{Single-Specialist Distillation}
\label{sec:exp_one_to_one}

\begin{table}[t]
  \centering
  \caption{\textbf{Single-specialist distillation:} each student is distilled from each modality's teacher separately. Base: before distillation; + HuluOPD: after the full recipe; green: gain over Base in points. Clin.: DentVLM\mbox{-}Clinical\mbox{-}Bench. Metrics as in \cref{sec:data}, $\times 100$; Avg.: over all VQA and VQA\mbox{-}MS items.}
  \label{tab:separate}
  \scriptsize
  \setlength{\tabcolsep}{0pt}
  \begin{tabular*}{\linewidth}{@{\extracolsep{\fill}}llccccc@{}}
  \toprule
   & & & \multicolumn{2}{c}{HuluMed\mbox{-}4B} & \multicolumn{2}{c}{Qwen3.5\mbox{-}0.8B} \\
  \cmidrule(l{4pt}){4-5}\cmidrule(l{4pt}){6-7}
  Benchmark & Task & Teacher & Base & + HuluOPD & Base & + HuluOPD \\
  \midrule
  \modrow{Ultrasound} \\
  \hspace{3pt}HuluOPDBench    & VQA & 73.44 & 47.51 & \g{64.00}{+16.49} & 45.80 & \g{60.31}{+14.51} \\
                              & Gen & 89.36 & 83.40 & \g{89.01}{+5.61} & 81.29 & \g{88.87}{+7.58} \\
  \hspace{3pt}OmniMedVQA      & VQA & 97.28 & 75.41 & \g{83.43}{+8.02} & 66.85 & \g{78.24}{+11.39} \\
  \hspace{3pt}U2\mbox{-}BENCH        & VQA & 58.07 & 40.83 & \g{52.28}{+11.45} & 40.46 & \g{50.59}{+10.13} \\
                              & Gen & 88.93 & 85.94 & \g{86.54}{+0.60} & 84.04 & \g{85.48}{+1.44} \\
  \midrule
  \modrow{Pathology} \\
  \hspace{3pt}HuluOPDBench    & VQA & 60.70 & 45.38 & \g{57.45}{+12.07} & 36.55 & \g{55.52}{+18.97} \\
                              & Gen & 84.43 & 81.80 & \g{84.24}{+2.44} & 81.20 & \g{84.00}{+2.80} \\
  \hspace{3pt}PathMMU         & VQA & 75.92 & 60.79 & \g{62.55}{+1.76} & 49.84 & \g{54.96}{+5.12} \\
  \hspace{3pt}PathVQA         & VQA & 69.12 & 46.36 & \g{60.28}{+13.92} & 43.75 & \g{58.49}{+14.74} \\
  \midrule
  \modrow{OCT} \\
  \hspace{3pt}HuluOPDBench    & VQA & 84.15 & 61.26 & \g{82.05}{+20.79} & 52.02 & \g{78.89}{+26.87} \\
                              & Gen & 86.71 & 81.68 & \g{83.68}{+2.00} & 80.50 & \g{82.40}{+1.90} \\
  \hspace{3pt}FunBench        & VQA & 91.56 & 72.95 & \g{90.45}{+17.50} & 75.33 & \g{89.10}{+13.77} \\
  \midrule
  \modrow{Fundus photography} \\
  \hspace{3pt}HuluOPDBench    & VQA & 67.27 & 49.02 & \g{61.25}{+12.23} & 39.34 & \g{59.99}{+20.65} \\
                              & Gen & 86.49 & 81.77 & \g{83.68}{+1.91} & 80.92 & \g{83.35}{+2.43} \\
  \hspace{3pt}FunBench        & VQA & 86.82 & 70.10 & \g{85.97}{+15.87} & 71.65 & \g{85.01}{+13.36} \\
                              & VQA\mbox{-}MS & 63.70 & 48.35 & \g{59.86}{+11.51} & 48.20 & \g{58.74}{+10.54} \\
  \midrule
  \modrow{Dental imaging} \\
  \hspace{3pt}DentVLM\mbox{-}Bench   & VQA & 77.79 & 52.32 & \g{72.11}{+19.79} & 43.98 & \g{69.65}{+25.67} \\
  \hspace{3pt}DentVLM\mbox{-}Clin.   & VQA & 76.62 & 47.25 & \g{70.14}{+22.89} & 50.50 & \g{68.15}{+17.65} \\
  \midrule
  \multicolumn{2}{@{}l}{Avg.} & 77.72 & 58.25 & \g{72.46}{+14.21} & 53.58 & \g{69.97}{+16.39} \\
  \bottomrule
  \end{tabular*}
\end{table}

We first distill each specialist into its own student, as the reference for the unified setting (\cref{tab:separate}).
HuluOPD improves both students on every benchmark: on question answering, HuluMed\mbox{-}4B gains 14.2 points on average, from 75\% to 93\% of the teachers' Avg., and Qwen3.5\mbox{-}0.8B, about $10\times$ smaller than the teachers, gains 16.4 points, from 69\% to 90\%, which closes 73\% and 68\% of the gap to the teachers.

These gains do not depend on our own teachers, on the choice of student, or on our own benchmark.
First, the largest gains (20.3 and 24.5 points) are on dental imaging, whose teacher is the released DentVLM, so the recipe works with an off-the-shelf specialist without modification.
Second, Qwen3.5\mbox{-}0.8B, a general-purpose model five times smaller than the medical HuluMed\mbox{-}4B, narrows its gap to HuluMed\mbox{-}4B from 4.7 to 2.5 points, and each student improves on every benchmark in the modalities where it must cross tokenizers.
Third, on the third-party benchmarks, which we did not build, the students reach the same 93\% and 90\% of the teachers' Avg. as on the full evaluation.

\subsection{Unified Multi-Specialist Distillation}
\label{sec:exp_five_to_one}

\makeatletter
\colorlet{tpGen}{Apricot!10}%
\colorlet{tpTea}{RoyalBlue!7}%
\colorlet{tpHulu}{Orchid!8}%
\colorlet{tpQwen}{TealBlue!8}%
\providecommand{\tpmark}[1]{\if@filesw\pdfsavepos\write\@auxout{\string\expandafter\string\gdef\string\csname\space tp@#1\string\endcsname{{\the\pdflastxpos}{\the\pdflastypos}}}\fi}
\providecommand{\tp@x}[1]{\expandafter\expandafter\expandafter\@firstoftwo\csname tp@#1\endcsname sp}
\providecommand{\tp@y}[1]{\expandafter\expandafter\expandafter\@secondoftwo\csname tp@#1\endcsname sp}
\providecommand{\tpabove}{7.9pt}%
\providecommand{\tpbelow}{3.5pt}%
\providecommand{\tpgap}{2pt}
\providecommand{\tp@mid}[2]{\dimexpr(\tp@x{#1}+\tp@x{#2})/2\relax}
\providecommand{\tp@ifmarks}[1]{\@tempswatrue\@for\tp@m:=o,T,B,gL,gR,tL,tR,hL,hR,qL,qR\do{\@ifundefined{tp@\tp@m}{\@tempswafalse}{}}\if@tempswa#1\fi}
\providecommand{\tpgroups}{%
  \tpdo{tpGen}{\tp@x{gL}-5pt}{\tp@mid{gR}{tL}-\tpgap/2}%
  \tpdo{tpTea}{\tp@mid{gR}{tL}+\tpgap/2}{\tp@mid{tR}{hL}-\tpgap/2}%
  \tpdo{tpHulu}{\tp@mid{tR}{hL}+\tpgap/2}{\tp@mid{hR}{qL}-\tpgap/2}%
  \tpdo{tpQwen}{\tp@mid{hR}{qL}+\tpgap/2}{\tp@x{qR}+1.3pt}}
\providecommand{\tppanel}[3]{%
  \tp@ifmarks{\rlap{\hskip\dimexpr#2-\tp@x{o}\relax
    \raisebox{\dimexpr\tp@y{B}-\tp@y{o}-\tpbelow\relax}[0pt][0pt]{\textcolor{#1}{\rule{\dimexpr#3-(#2)\relax}{\dimexpr\tp@y{T}-\tp@y{B}+\tpabove+\tpbelow\relax}}}}}}
\providecommand{\tpband}[3]{%
  \tp@ifmarks{\rlap{\hskip\dimexpr#2-\tp@x{o}\relax\textcolor{#1!93!black}{\rule[-2.2pt]{\dimexpr#3-(#2)\relax}{8.2pt}}}}}
\providecommand{\modrowS}[2]{\multicolumn{#1}{@{}l@{}}{\makebox[0pt][l]{\setlength{\fboxsep}{0pt}\colorbox{black!7}{\makebox[\linewidth][l]{\rule[-2.2pt]{0pt}{8.2pt}\hspace{1pt}\textit{#2}}}}\let\tpdo\tpband\tpgroups}}
\makeatother
\begin{table*}[t]
  \centering
  \caption{\textbf{Unified multi-specialist distillation:} one student is distilled from all five teachers. Open Generalists: released medical VLMs, evaluated as released under the same protocol. Bold: best among generalists and students. Otherwise as in \cref{tab:separate}.}
  \label{tab:unified}
  \scriptsize
  \setlength{\tabcolsep}{0pt}
  \setlength{\genw}{0pt}\genwmax{HealthGPT}\genwmax{MedGemma}\genwmax{Pro-4B~[00]}\genwmax{1.5-4B~[00]}\genwmax{30A3~[00]}\genwmax{HuluMed}\genwmax{00.00}
  \leavevmode\tpmark{o}%
  \let\tpdo\tppanel\tpgroups%
  \begin{tabular*}{\linewidth}{@{\extracolsep{\fill}}lllccccccccc@{}}
  \toprule
   \tpmark{T} & & & \multicolumn{4}{c}{Open Generalists} \\[0.5pt]%
   & & & \multicolumn{4}{c}{\genrule} & & \multicolumn{2}{c}{HuluMed\mbox{-}4B} & \multicolumn{2}{c}{Qwen3.5\mbox{-}0.8B} \\
  \cmidrule{9-10}\cmidrule{11-12}
  Modality & Benchmark & Task & \tpmark{gL}\gh{HealthGPT}{Pro\mbox{-}4B~\cite{lin2025healthgpt}} & \gh{MedGemma}{1.5\mbox{-}4B~\cite{sellergren2026medgemma15}} & \gh{Lingshu}{7B~\cite{xu2025lingshu}} & \gh{HuluMed}{30A3~\cite{jiang2025hulumed}}\tpmark{gR} & \tpmark{tL}Teacher\tpmark{tR} & Base & + HuluOPD & Base & + HuluOPD \\
  \midrule
  Ultrasound & HuluOPDBench    & VQA & 49.07 & 42.46 & 47.27 & 52.72 & 73.44 & 47.51 & \g{\textbf{60.65}}{+13.14} & 45.80 & \g{58.03}{+12.23} \\
  & & Gen & 83.19 & 81.96 & 84.04 & 83.15 & 89.36 & 83.40 & \g{\textbf{88.70}}{+5.30} & 81.29 & \g{88.14}{+6.85} \\
  & OmniMedVQA      & VQA & 74.72 & 61.03 & 68.12 & 82.09 & 97.28 & 75.41 & \g{\textbf{84.79}}{+9.38} & 66.85 & \g{76.35}{+9.50} \\
  & U2\mbox{-}BENCH        & VQA & 41.30 & 35.45 & 43.48 & 49.67 & 58.07 & 40.83 & \g{\textbf{52.17}}{+11.34} & 40.46 & \g{47.31}{+6.85} \\
  & & Gen & 85.48 & 84.17 & 84.83 & 84.95 & 88.93 & 85.94 & \g{\textbf{86.74}}{+0.80} & 84.04 & \g{86.65}{+2.61} \\
  \midrule
  Pathology & HuluOPDBench    & VQA & 42.31 & 38.33 & 37.35 & \textbf{59.97} & 60.70 & 45.38 & \g{56.66}{+11.28} & 36.55 & \g{52.21}{+15.66} \\
  & & Gen & 83.82 & 81.85 & 82.04 & 81.97 & 84.43 & 81.80 & \g{\textbf{84.18}}{+2.38} & 81.20 & \g{83.89}{+2.69} \\
  & PathMMU         & VQA & 62.34 & 46.59 & 52.66 & \textbf{70.86} & 75.92 & 60.79 & \g{62.72}{+1.93} & 49.84 & \g{52.77}{+2.93} \\
  & PathVQA         & VQA & \textbf{63.60} & 39.80 & 56.28 & 61.99 & 69.12 & 46.36 & \g{58.67}{+12.31} & 43.75 & \g{54.16}{+10.41} \\
  \midrule
  OCT & HuluOPDBench    & VQA & 46.58 & 28.84 & 39.86 & 75.00 & 84.15 & 61.26 & \g{\textbf{82.39}}{+21.13} & 52.02 & \g{78.42}{+26.40} \\
  & & Gen & 81.49 & 80.81 & 82.77 & 81.47 & 86.71 & 81.68 & \g{\textbf{83.49}}{+1.81} & 80.50 & \g{82.83}{+2.33} \\
  & FunBench        & VQA & 69.33 & 50.29 & 65.80 & 82.29 & 91.56 & 72.95 & \g{\textbf{91.32}}{+18.37} & 75.33 & \g{89.41}{+14.08} \\
  \midrule
  Fundus photography & HuluOPDBench    & VQA & 45.97 & 49.81 & 45.07 & 60.42 & 67.27 & 49.02 & \g{\textbf{62.75}}{+13.73} & 39.34 & \g{60.23}{+20.89} \\
  & & Gen & 82.28 & 81.94 & 83.35 & 82.18 & 86.49 & 81.77 & \g{\textbf{83.77}}{+2.00} & 80.92 & \g{83.33}{+2.41} \\
  & FunBench        & VQA & 67.81 & 71.36 & 61.11 & 82.62 & 86.82 & 70.10 & \g{\textbf{86.63}}{+16.53} & 71.65 & \g{85.40}{+13.75} \\
  & & VQA\mbox{-}MS & 50.28 & 55.38 & 41.91 & 59.60 & 63.70 & 48.35 & \g{\textbf{63.16}}{+14.81} & 48.20 & \g{58.69}{+10.49} \\
  \midrule
  Dental imaging & DentVLM\mbox{-}Bench   & VQA & 48.20 & 21.97 & 55.09 & 51.90 & 77.79 & 52.32 & \g{\textbf{73.79}}{+21.47} & 43.98 & \g{70.30}{+26.32} \\
  & DentVLM\mbox{-}Clin.   & VQA & 44.48 & 27.02 & 51.34 & 47.12 & 76.62 & 47.25 & \g{\textbf{71.82}}{+24.57} & 50.50 & \g{67.95}{+17.45} \\
  \midrule
  \multicolumn{3}{@{}l}{\tpmark{B}Avg.} & 56.12 & 46.15 & 52.95 & 67.51 & 77.72 & \tpmark{hL}58.25 & \g{\textbf{72.87}}{+14.62}\tpmark{hR} & \tpmark{qL}53.58 & \g{68.94}{+15.36}\tpmark{qR} \\
  \bottomrule
  \end{tabular*}
\end{table*}
\begin{figure*}[t]
  \centering
  \includegraphics[width=\linewidth]{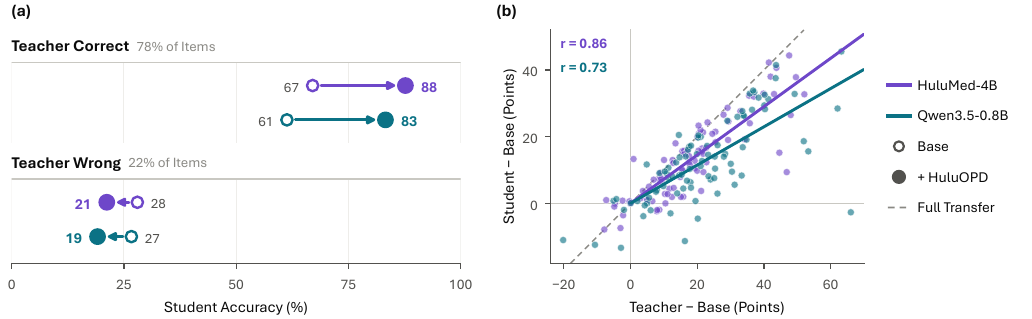}
  \caption{\textbf{What the students take from their teachers.} (a) Accuracy before (hollow) and after HuluOPD (filled), on items the teacher gets right (97,001) and wrong (27,348). (b) For each of the 87 source datasets with at least 100 items, the student's gain over its base model against the teacher's lead over that model. Solid lines: least-squares fits through the origin; dashed line: student equals teacher.}
  \label{fig:transfer_lead}
\end{figure*}

We then distill all five specialists into one student at once (\cref{tab:unified}): it is warmed up on 80k of the single-specialist warm-up responses and then distilled for two epochs over 344k HuluOPDCorpus prompts, 38\% of the corpus (Appendices~\ref{sec:supp_data} and~\ref{sec:supp_impl}).
HuluOPD again improves both students on every benchmark: on question answering, unified HuluMed\mbox{-}4B gains 14.6 points and reaches 94\% of the five teachers' Avg.\ with 10\% of their combined parameters, and unified Qwen3.5\mbox{-}0.8B gains 15.4 points and reaches 89\% with 2\%.
The gains cover all four task groups (Appendix~\cref{tab:taskgroup}).
On diagnosis and grading, recognition, and detection and localization, the unified students reach 94\%, 92\%, and 98\% (HuluMed\mbox{-}4B) and 89\%, 86\%, and 96\% (Qwen3.5\mbox{-}0.8B) of the teachers' score; on generation, they end 1.2 and 1.6 BERTScore points below the teachers.

Consolidating five specialists into one student costs almost nothing.
With one model in place of five and 38\% of the prompts, unified HuluMed\mbox{-}4B is 0.41 points above the single-specialist students of \cref{tab:separate} on average, and unified Qwen3.5\mbox{-}0.8B keeps 94\% of their gain.
Both match or exceed them on OCT, fundus photography, and dental imaging, and trail mainly on ultrasound and pathology, the two modalities whose prompts shrink by three to four times in the unified setting.

The unified students also outperform much larger open generalists.
On the Avg., both students are ahead of all four evaluated generalists: unified HuluMed\mbox{-}4B exceeds HuluMed\mbox{-}30A3, a model of its own family with seven times its parameters, by 5.4 points, and unified Qwen3.5\mbox{-}0.8B, with 36 times fewer parameters, exceeds it by 1.4 points.
This lead holds benchmark by benchmark: before distillation, neither student led the best generalist on any question-answering benchmark, whereas unified HuluMed\mbox{-}4B now leads on 10 of the 13 and on all five generation benchmarks, most clearly on dental imaging (by 19.0 points on average).

\subsection{Analysis}
\label{sec:analysis}

Beyond the gains in absolute scores, we ask what the unified students of \cref{tab:unified} actually learn from their teachers.

\paragraph{The students learn faithfully from their teachers.}
On the items the teacher answers correctly, both students improve sharply (\cref{fig:transfer_lead}a), and they take over not only the teacher's answer but also its reasoning (\cref{fig:cases}). In the example shown, the base models deny the key findings and misdiagnose the image, whereas after distillation both students report the same findings as the teacher and reason from them to the correct diagnosis. On the items the teacher gets wrong, however, the students' accuracy drops. When teacher and student are both wrong, the students now pick the teacher's wrong option more often (72\% and 65\% of cases, against 55\% and 44\% before). With no labels to correct them, the students thus follow the teacher faithfully, inheriting its errors along with its knowledge. Yet this negative transfer is much weaker than the positive one: the drop is only about a third of the gain and affects far fewer items, so overall the students gain roughly ten times as much as they lose: what they inherit is mostly the teacher's knowledge, not its mistakes.

\providecommand{\cmark}{\textcolor{ForestGreen}{$\checkmark$}}
\providecommand{\xmark}{\textcolor{BrickRed}{$\times$}}
\begin{figure}[t]
  \scriptsize
  \newcommand{\fnd}[1]{{\setlength{\fboxsep}{0pt}\colorbox{Goldenrod!80}{\vphantom{hg}\hspace{0.6pt}\textbf{#1}\hspace{0.6pt}}}}
  \newcommand{\den}[1]{{\setlength{\fboxsep}{0pt}\colorbox{BrickRed!42}{\vphantom{hg}\hspace{0.6pt}\textbf{#1}\hspace{0.6pt}}}}
  \newcommand{\caserow}[2]{\noindent{\setlength{\fboxsep}{2.5pt}\colorbox{#1}{\parbox{\dimexpr\linewidth-5pt\relax}{\raggedright #2}}}\par\vspace{2pt}}
  \noindent\begin{minipage}[t]{0.27\linewidth}\vspace{0pt}\includegraphics[width=\linewidth]{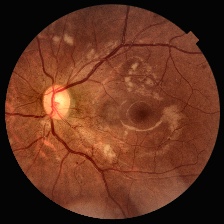}\end{minipage}\hfill
  \begin{minipage}[t]{0.70\linewidth}\vspace{0pt}\raggedright
    \textbf{Question (abridged).} Select the most appropriate diagnosis for this fundus image: A.~Wet age-related macular degeneration; B.~Normal; C.~Dry age-related macular degeneration; D.~Diabetic retinopathy; E.~Central serous chorioretinopathy; F.~Glaucoma. Explain your reasoning and then give the final answer.\par\vspace{3pt}
    \textbf{Reference.} D.~Diabetic retinopathy
  \end{minipage}\par\vspace{5pt}
  \caserow{RoyalBlue!13}{\textbf{Teacher} \cmark\ \emph{D.~Diabetic retinopathy}: ``The fundus photograph reveals multiple yellowish-white deposits consistent with \fnd{hard exudates} scattered throughout the posterior pole [\ldots] There are also several small, round, red lesions indicative of \fnd{microaneurysms} or \fnd{dot-and-blot hemorrhages} [\ldots]''}
  \vspace{2pt}
  \caserow{BrickRed!11}{\textbf{HuluMed\mbox{-}4B, Base} \xmark\ \emph{B.~Normal}: ``[\ldots] There are \den{no hemorrhages}, \den{microaneurysms}, or \den{exudates} typically associated with \mbox{diabetic retinopathy [\ldots]''}}
  \caserow{ForestGreen!14}{\textbf{HuluMed\mbox{-}4B, + HuluOPD} \cmark\ \emph{D.~Diabetic retinopathy}: ``The fundus photograph reveals multiple yellowish-white \fnd{hard exudates} arranged in a circinate pattern temporal to the fovea, accompanied by scattered \fnd{dot-and-blot hemorrhages} in the posterior pole. [\ldots] There is evidence of \fnd{microaneurysms} [\ldots]''}
  \vspace{2pt}
  \caserow{BrickRed!11}{\textbf{Qwen3.5\mbox{-}0.8B, Base} \xmark\ \emph{C.~Dry age-related macular degeneration}: ``[\ldots] D. Diabetic retinopathy: Unlikely --- \den{no hemorrhages}, \den{microaneurysms}, or \mbox{\den{exudates}. [\ldots]''}}
  \caserow{ForestGreen!14}{\textbf{Qwen3.5\mbox{-}0.8B, + HuluOPD} \cmark\ \emph{D.~Diabetic retinopathy}: ``The fundus photograph reveals multiple yellowish-white deposits consistent with \fnd{hard exudates} scattered in the posterior pole [\ldots] There are also several small, round, red lesions [\ldots] which represent \fnd{microaneurysms} or \fnd{dot-and-blot hemorrhages}. [\ldots]''}
  \vspace{-2pt}
  \caption{\textbf{Case study:} a typical HuluOPDBench item. {\setlength{\fboxsep}{1pt}\colorbox{Goldenrod!80}{Yellow}} marks the key findings where a model reports them, {\setlength{\fboxsep}{1pt}\colorbox{BrickRed!42}{dark red}} where it denies them; [\ldots]: omitted text.}
  \label{fig:cases}
\end{figure}

\paragraph{The gain follows the teacher's lead.}
How much a student gains depends on how far its teacher is ahead of it (\cref{fig:transfer_lead}b).
Across datasets, the larger the teacher's lead over the base model, the larger the student's gain, and where the teacher is not ahead, the gain disappears.
For every point of the teacher's lead, HuluMed\mbox{-}4B gains about 0.7 points and the five-times-smaller Qwen3.5\mbox{-}0.8B about 0.6, so a larger student learns more from the same teachers.
After distillation, most of the students' remaining errors are on items that the teacher also gets wrong (64\% and 58\%, against 38\% and 35\% before).
What limits the students is therefore mainly the teachers themselves, and a stronger specialist should yield a stronger student.

\providecommand{\cmark}{\textcolor{ForestGreen}{$\checkmark$}}
\providecommand{\xmark}{\textcolor{BrickRed}{$\times$}}

\subsection{Comparison and Ablation}
\label{sec:ablation}

\providecommand{\cmark}{\textcolor{ForestGreen}{$\checkmark$}}
\providecommand{\xmark}{\textcolor{BrickRed}{$\times$}}

\paragraph{Comparison with OPD methods.}
HuluOPD is the only OPD method that distills several heterogeneous teachers into one VLM (\cref{tab:capability}): prior cross-tokenizer methods distill a single teacher into a text-only model, and prior multi-teacher methods need teachers with the student's tokenizer.

\paragraph{Loss ablation.}
Since no existing method covers this setting as a whole, we keep the recipe fixed and vary its loss in four variants, A--D, each changing one component of the one before (\cref{tab:ablation}).
Variant~A pairs the $i$-th student token with the $i$-th teacher token and applies only the sorted-probability distance $W$ of \cref{eq:hybrid} to the full vocabularies, as in ULD~\cite{boizard2024uld} without its cross-entropy term.
Variant~B replaces it with the hybrid loss, matching candidates by a fixed map between identical token strings~\cite{gold2025}.
Variant~C aligns by text instead of by position (\cref{sec:loss}).
Variant~D, our full loss, pairs candidates by continuation instead of the fixed map (\cref{eq:pairing}).
Each variant distills HuluMed\mbox{-}4B from the unified warm-up checkpoint on 50k prompts, about 15\% of the unified prompts (Appendix~\ref{sec:supp_ablation}).

\suppressfloats[t]
\begin{table}[t]
  \centering
  \scriptsize
  \caption{\textbf{Comparison with representative distillation methods.} Multi: several teachers; Hetero.: teachers may differ from the student in architecture and tokenizer; Label-free: no answers.}
  \label{tab:capability}
  \setlength{\tabcolsep}{2pt}
  \begin{tabular*}{\linewidth}{@{}l@{\extracolsep{\fill}}cccccc@{}}
  \toprule
  Method & On-policy & Multi & Hetero. & VLM & Medical & Label-free \\
  \midrule
  ULD~\cite{boizard2024uld}          & \xmark & \xmark & \cmark & \xmark & \xmark & \cmark \\
  GKD~\cite{agarwal2024gkd}          & \cmark & \xmark & \xmark & \xmark & \xmark & \cmark \\
  GOLD~\cite{gold2025}               & \cmark & \xmark & \cmark & \xmark & \xmark & \cmark \\
  Med\mbox{-}OPD~\cite{qian2026medopd}      & \cmark & \xmark & \xmark & \cmark & \cmark & \xmark \\
  MOPD~\cite{ma2026mopd}             & \cmark & \cmark & \xmark & \xmark & \xmark & \cmark \\
  OPOD~\cite{zhao2026opod}           & \cmark & \cmark & \xmark & \cmark & \xmark & \xmark \\
  \midrule
  HuluOPD (ours)                     & \cmark & \cmark & \cmark & \cmark & \cmark & \cmark \\
  \bottomrule
  \end{tabular*}

  \vspace{8pt}
  \caption{\textbf{Loss ablation} (HuluMed\mbox{-}4B, unified setting; scores as the Avg.\ of \cref{tab:separate}). Shared / Unshared: teacher and student share a base vocabulary (OCT, fundus photography, dental imaging) or not (ultrasound, pathology). Red: lowered by repetition.}
  \label{tab:ablation}
  \setlength{\tabcolsep}{0pt}
  \begin{tabular*}{\linewidth}{@{}l@{\hspace{7pt}}@{\extracolsep{\fill}}lllccc@{}}
  \toprule
  & Alignment & Loss & Pairing & Shared & Unshared & Avg. \\
  \midrule
  A & Position & Sorted & --           & 75.34 & 60.68 & 68.92 \\
  B & Position & Hybrid & Fixed        & 78.44 & \textcolor{BrickRed}{58.15} & 69.56 \\
  C & Text     & Hybrid & Fixed        & 78.40 & 61.47 & 70.99 \\
  D (ours) & Text     & Hybrid & Continuation & \textbf{78.46} & \textbf{61.71} & \textbf{71.13} \\
  \bottomrule
  \end{tabular*}
\end{table}

Even in this light setting, every variant improves the Avg.\ over the one before, and D is the best in every column (\cref{tab:ablation}).
Compared with A, B gains 3.1 points on the shared-vocabulary modalities: every candidate is matched there, so the hybrid loss compares each token with its counterpart and not only the shapes of the two distributions.
On the unshared-vocabulary modalities, however, B falls 2.5 points below A (red): its positional pairing misaligns the two tokenizations, and this misaligned supervision leads to repetitive outputs on pathology (Appendix~\cref{fig:ablation_examples}).
Compared with B, C aligns the tokenizations by text, which removes this failure and adds 3.3 points on these modalities.
Compared with C, D pairs the candidates by continuation, which adds a further 0.2 points.
On the shared-vocabulary modalities, where alignment and pairing nearly coincide, B, C, and D agree within 0.06 points pooled and 0.25 points on any one modality, in either direction (Appendix~\ref{sec:supp_ablation}).

\section{Conclusion}
\label{sec:conclusion}

We presented HuluOPD, an on-policy distillation recipe that unifies heterogeneous medical specialists into one lightweight VLM using only unlabeled prompts, instead of deploying one large model per specialty or pooling labeled data into a generalist.
The student answers each prompt itself, the specialist of the prompt's modality supervises its response, and a continuation-aligned cross-tokenizer loss pairs their predictions by the text they complete.
Distilled on HuluOPDCorpus and scored on 132k test items including HuluOPDBench, a single student with 10\% of the five specialists' combined parameters retains 94\% of their question-answering performance, and one with 2\% retains 89\%; both outperform, on average, four such generalists with up to 36 times their parameters.
A student's gain grows with its teacher's lead, so stronger specialists should yield stronger students.
To our knowledge, this is the first demonstration that heterogeneous medical specialist VLMs can be unified into one lightweight model without requiring their training data or a shared architecture, which opens a feasible path past the privacy barriers and fragmentation of medical AI toward more transparent general medical intelligence.

\section{Limitations}
\label{sec:limitations}
Four of our five teachers are our own stand-ins for leading specialists, and their training data share public sources with our prompts, so more released specialists and an institution's own prompts remain untested.
We cover only 2D images, leaving 3D and temporal data to future work, and on-policy training costs more than off-policy training.

{
    \small
    \bibliographystyle{ieeenat_fullname}
    \bibliography{main}

\begin{thebibliography}{99}
\providecommand{\natexlab}[1]{#1}
\providecommand{\url}[1]{\texttt{#1}}
\expandafter\ifx\csname urlstyle\endcsname\relax
  \providecommand{\doi}[1]{doi: #1}\else
  \providecommand{\doi}{doi: \begingroup \urlstyle{rm}\Url}\fi

\bibitem[Abr{\`a}moff et~al.(2013)Abr{\`a}moff, Folk, Han, Walker, Williams, Russell, Massin, Cochener, Gain, Tang, et~al.]{abramoff2013messidor2}
Michael~D. Abr{\`a}moff, James~C. Folk, Dennis~P. Han, Jonathan~D. Walker, David~F. Williams, Stephen~R. Russell, Pascale Massin, Beatrice Cochener, Philippe Gain, Li Tang, et~al.
\newblock Automated analysis of retinal images for detection of referable diabetic retinopathy.
\newblock \emph{JAMA Ophthalmology}, 131\penalty0 (3), 2013.

\bibitem[Ahn et~al.(2018)Ahn, Kim, Ahn, Cho, Lee, and Kim]{ahn2018glaucoma}
Jin~Mo Ahn, Sangsoo Kim, Kwang-Sung Ahn, Sung-Hoon Cho, Kwan~Bok Lee, and Ungsoo~Samuel Kim.
\newblock A deep learning model for the detection of both advanced and early glaucoma using fundus photography.
\newblock \emph{PLOS ONE}, 13\penalty0 (11):\penalty0 e0207982, 2018.

\bibitem[Al-Dhabyani et~al.(2020)Al-Dhabyani, Gomaa, Khaled, and Fahmy]{aldhabyani2020busi}
Walid Al-Dhabyani, Mohammed Gomaa, Hussien Khaled, and Aly Fahmy.
\newblock Dataset of breast ultrasound images.
\newblock \emph{Data in Brief}, 28:\penalty0 104863, 2020.

\bibitem[{Anagha Choudhari}(2022)]{choudhari2022pcos}
{Anagha Choudhari}.
\newblock {PCOS} detection using ultrasound images.
\newblock Kaggle, \url{https://www.kaggle.com/datasets/anaghachoudhari/pcos-detection-using-ultrasound-images}, 2022.

\bibitem[Aresta et~al.(2019)Aresta, Ara{\'u}jo, Kwok, Chennamsetty, Safwan, Alex, Marami, Prastawa, Chan, Donovan, et~al.]{aresta2019bach}
Guilherme Aresta, Teresa Ara{\'u}jo, Scotty Kwok, Sai~Saketh Chennamsetty, Mohammed Safwan, Varghese Alex, Bahram Marami, Marcel Prastawa, Monica Chan, Michael Donovan, et~al.
\newblock {BACH}: Grand challenge on breast cancer histology images.
\newblock \emph{Medical Image Analysis}, 56:\penalty0 122--139, 2019.

\bibitem[Arunachalam et~al.(2019)Arunachalam, Mishra, Daescu, Cederberg, Rakheja, Sengupta, Leonard, Hallac, and Leavey]{arunachalam2019osteosarcoma}
Harish~Babu Arunachalam, Rashika Mishra, Ovidiu Daescu, Kevin Cederberg, Dinesh Rakheja, Anita Sengupta, David Leonard, Rami Hallac, and Patrick Leavey.
\newblock Viable and necrotic tumor assessment from whole slide images of osteosarcoma using machine-learning and deep-learning models.
\newblock \emph{PLOS ONE}, 14\penalty0 (4):\penalty0 e0210706, 2019.

\bibitem[{Azouz Maroua}(2022)]{maroua2022auitd}
{Azouz Maroua}.
\newblock Algerian ultrasound images thyroid dataset: {AUITD}.
\newblock Kaggle, \url{https://www.kaggle.com/datasets/azouzmaroua/algeria-ultrasound-images-thyroid-dataset-auitd}, 2022.

\bibitem[Beltagy et~al.(2020)Beltagy, Peters, and Cohan]{beltagy2020longformer}
Iz Beltagy, Matthew~E. Peters, and Arman Cohan.
\newblock Longformer: The long-document transformer.
\newblock \emph{arXiv preprint arXiv:2004.05150}, 2020.

\bibitem[Bogunovic et~al.(2019)Bogunovic, Venhuizen, Klimscha, Apostolopoulos, Bab-Hadiashar, Bagci, Beg, Bekalo, Chen, Ciller, et~al.]{bogunovic2019retouch}
Hrvoje Bogunovic, Freerk Venhuizen, Sophie Klimscha, Stefanos Apostolopoulos, Alireza Bab-Hadiashar, Ulas Bagci, Mirza~Faisal Beg, Loza Bekalo, Qiang Chen, Carlos Ciller, et~al.
\newblock {RETOUCH}: The retinal {OCT} fluid detection and segmentation benchmark and challenge.
\newblock \emph{IEEE Transactions on Medical Imaging}, 38\penalty0 (8):\penalty0 1858--1874, 2019.

\bibitem[Bolhasani et~al.(2020)Bolhasani, Amjadi, Tabatabaeian, and Jassbi]{bolhasani2020databiox}
Hamidreza Bolhasani, Elham Amjadi, Maryam Tabatabaeian, and Somayyeh~Jafarali Jassbi.
\newblock A histopathological image dataset for grading breast invasive ductal carcinomas.
\newblock \emph{Informatics in Medicine Unlocked}, 19:\penalty0 100341, 2020.

\bibitem[Borkowski et~al.(2019)Borkowski, Bui, Thomas, Wilson, DeLand, and Mastorides]{borkowski2019lc25000}
Andrew~A. Borkowski, Marilyn~M. Bui, L.~Brannon Thomas, Catherine~P. Wilson, Lauren~A. DeLand, and Stephen~M. Mastorides.
\newblock Lung and colon cancer histopathological image dataset ({LC25000}).
\newblock \emph{arXiv preprint arXiv:1912.12142}, 2019.

\bibitem[Born et~al.(2021)Born, Wiedemann, Cossio, Buhre, Br{\"a}ndle, Leidermann, Goulet, Aujayeb, Moor, Rieck, and Borgwardt]{born2021pocus}
Jannis Born, Nina Wiedemann, Manuel Cossio, Charlotte Buhre, Gabriel Br{\"a}ndle, Konstantin Leidermann, Julie Goulet, Avinash Aujayeb, Michael Moor, Bastian Rieck, and Karsten Borgwardt.
\newblock Accelerating detection of lung pathologies with explainable ultrasound image analysis.
\newblock \emph{Applied Sciences}, 11\penalty0 (2):\penalty0 672, 2021.

\bibitem[Brummer et~al.(2023)Brummer, P{\"o}l{\"o}nen, Mustjoki, and Br{\"u}ck]{brummer2023renalcell}
Otso Brummer, Petri P{\"o}l{\"o}nen, Satu Mustjoki, and Oscar Br{\"u}ck.
\newblock Computational textural mapping harmonises sampling variation and reveals multidimensional histopathological fingerprints.
\newblock \emph{British Journal of Cancer}, 129\penalty0 (4):\penalty0 683--695, 2023.

\bibitem[Burgos-Artizzu et~al.(2020)Burgos-Artizzu, Coronado-Guti{\'e}rrez, Valenzuela-Alcaraz, Bonet-Carne, Eixarch, Crispi, and Gratac{\'o}s]{burgosartizzu2020fetalplanes}
Xavier~P. Burgos-Artizzu, David Coronado-Guti{\'e}rrez, Brenda Valenzuela-Alcaraz, Elisenda Bonet-Carne, Elisenda Eixarch, Fatima Crispi, and Eduard Gratac{\'o}s.
\newblock Evaluation of deep convolutional neural networks for automatic classification of common maternal fetal ultrasound planes.
\newblock \emph{Scientific Reports}, 10\penalty0 (1):\penalty0 10200, 2020.

\bibitem[Cen et~al.(2021)Cen, Ji, Lin, Ju, Lin, Li, Wang, Yang, Liu, Tan, et~al.]{cen2021jsiec}
Ling-Ping Cen, Jie Ji, Jian-Wei Lin, Si-Tong Ju, Hong-Jie Lin, Tai-Ping Li, Yun Wang, Jian-Feng Yang, Yu-Fen Liu, Shaoying Tan, et~al.
\newblock Automatic detection of 39 fundus diseases and conditions in retinal photographs using deep neural networks.
\newblock \emph{Nature Communications}, 12\penalty0 (1):\penalty0 4828, 2021.

\bibitem[Chen(2017)]{chen2017retinadataset}
Yiwei Chen.
\newblock retina\_dataset.
\newblock GitHub repository, \url{https://github.com/yiweichen04/retina_dataset}, 2017.

\bibitem[Da et~al.(2022)Da, Huang, Li, Zuo, Zhang, Liu, Chen, Li, Xu, Hu, et~al.]{da2022digestpath}
Qian Da, Xiaodi Huang, Zhongyu Li, Yanfei Zuo, Chenbin Zhang, Jingxin Liu, Wen Chen, Jiahui Li, Dou Xu, Zhiqiang Hu, et~al.
\newblock {DigestPath}: A benchmark dataset with challenge review for the pathological detection and segmentation of digestive-system.
\newblock \emph{Medical Image Analysis}, 80:\penalty0 102485, 2022.

\bibitem[Decenci{\`e}re et~al.(2013)Decenci{\`e}re, Cazuguel, Zhang, Thibault, Klein, Meyer, Marcotegui, Quellec, Lamard, Danno, et~al.]{decenciere2013eophtha}
E. Decenci{\`e}re, G. Cazuguel, X. Zhang, G. Thibault, J.-C. Klein, F. Meyer, B. Marcotegui, G. Quellec, M. Lamard, R. Danno, et~al.
\newblock {TeleOphta}: Machine learning and image processing methods for teleophthalmology.
\newblock \emph{IRBM}, 34\penalty0 (2):\penalty0 196--203, 2013.

\bibitem[Decenci{\`e}re et~al.(2014)Decenci{\`e}re, Zhang, Cazuguel, Lay, Cochener, Trone, Gain, Ordonez, Massin, Erginay, Charton, and Klein]{decenciere2014messidor}
Etienne Decenci{\`e}re, Xiwei Zhang, Guy Cazuguel, Bruno Lay, B{\'e}atrice Cochener, Caroline Trone, Philippe Gain, Richard Ordonez, Pascale Massin, Ali Erginay, B{\'e}atrice Charton, and Jean-Claude Klein.
\newblock Feedback on a publicly distributed image database: The {Messidor} database.
\newblock \emph{Image Analysis \& Stereology}, 33\penalty0 (3), 2014.

\bibitem[Diaz-Pinto et~al.(2019)Diaz-Pinto, Morales, Naranjo, K{\"o}hler, Mossi, and Navea]{diazpinto2019acrima}
Andres Diaz-Pinto, Sandra Morales, Valery Naranjo, Thomas K{\"o}hler, Jose~M. Mossi, and Amparo Navea.
\newblock {CNNs} for automatic glaucoma assessment using fundus images: an extensive validation.
\newblock \emph{BioMedical Engineering OnLine}, 18\penalty0 (1):\penalty0 29, 2019.

\bibitem[Fang et~al.(2024)Fang, Li, Wu, Fu, Sun, Orlando, Bogunovi{\'c}, Zhang, and Xu]{fang2024palm}
Huihui Fang, Fei Li, Junde Wu, Huazhu Fu, Xu Sun, Jos{\'e}~Ignacio Orlando, Hrvoje Bogunovi{\'c}, Xiulan Zhang, and Yanwu Xu.
\newblock Open fundus photograph dataset with pathologic myopia recognition and anatomical structure annotation.
\newblock \emph{Scientific Data}, 11\penalty0 (1):\penalty0 99, 2024.

\bibitem[Fumero~Batista et~al.(2020)Fumero~Batista, Diaz-Aleman, Sigut, Alayon, Arnay, and Angel-Pereira]{fumerobatista2020rimonedl}
Francisco~Jos{\'e} Fumero~Batista, Tinguaro Diaz-Aleman, Jose Sigut, Silvia Alayon, Rafael Arnay, and Denisse Angel-Pereira.
\newblock {RIM}-{ONE} {DL}: A unified retinal image database for assessing glaucoma using deep learning.
\newblock \emph{Image Analysis \& Stereology}, 39\penalty0 (3):\penalty0 161--167, 2020.

\bibitem[Gamper and Rajpoot(2021)]{gamper2021arch}
Jevgenij Gamper and Nasir Rajpoot.
\newblock Multiple instance captioning: Learning representations from histopathology textbooks and articles.
\newblock In \emph{CVPR}, 2021.

\bibitem[Gholami et~al.(2020)Gholami, Roy, Parthasarathy, and Lakshminarayanan]{gholami2020octid}
Peyman Gholami, Priyanka Roy, Mohana~Kuppuswamy Parthasarathy, and Vasudevan Lakshminarayanan.
\newblock {OCTID}: Optical coherence tomography image database.
\newblock \emph{Computers \& Electrical Engineering}, 81:\penalty0 106532, 2020.

\bibitem[G{\'o}mez-Flores et~al.(2024)G{\'o}mez-Flores, Gregorio-Calas, and Coelho~de Albuquerque~Pereira]{gomezflores2024busbra}
Wilfrido G{\'o}mez-Flores, Maria~Julia Gregorio-Calas, and Wagner Coelho~de Albuquerque~Pereira.
\newblock {BUS}-{BRA}: A breast ultrasound dataset for assessing computer-aided diagnosis systems.
\newblock \emph{Medical Physics}, 51\penalty0 (4):\penalty0 3110--3123, 2024.

\bibitem[Gong et~al.(2023)Gong, Chen, Chen, Li, Li, and Chen]{gong2023tn3k}
Haifan Gong, Jiaxin Chen, Guanqi Chen, Haofeng Li, Guanbin Li, and Fei Chen.
\newblock Thyroid region prior guided attention for ultrasound segmentation of thyroid nodules.
\newblock \emph{Computers in Biology and Medicine}, 155:\penalty0 106389, 2023.

\bibitem[Han et~al.(2022)Han, Pan, Yan, Lin, Li, Yao, Lv, Shi, Mai, Lin, et~al.]{han2022wsss4luad}
Chu Han, Xipeng Pan, Lixu Yan, Huan Lin, Bingbing Li, Su Yao, Shanshan Lv, Zhenwei Shi, Jinhai Mai, Jiatai Lin, et~al.
\newblock {WSSS4LUAD}: Grand challenge on weakly-supervised tissue semantic segmentation for lung adenocarcinoma.
\newblock \emph{arXiv preprint arXiv:2204.06455}, 2022.

\bibitem[Handa et~al.(2025)Handa, Saini, Dutta, Pathak, Choudhary, Goel, and Dhanao]{handa2025pcosgen}
Palak Handa, Anushka Saini, Siddhant Dutta, Harsh Pathak, Nishi Choudhary, Nidhi Goel, and Jasdeep~Kaur Dhanao.
\newblock {PCOSGen}-train dataset.
\newblock Zenodo, \url{https://doi.org/10.5281/zenodo.14592001}, 2025.

\bibitem[He et~al.(2020)He, Zhang, Mou, Xing, and Xie]{he2020pathvqa}
Xuehai He, Yichen Zhang, Luntian Mou, Eric Xing, and Pengtao Xie.
\newblock {PathVQA}: 30000+ questions for medical visual question answering.
\newblock \emph{arXiv preprint arXiv:2003.10286}, 2020.

\bibitem[Hou et~al.(2024)Hou, Hua, Zhang, Ji, Zhang, Jiang, Li, Wu, Zhao, Sun, Cao, and Wang]{hou2024thyroidus}
Xiaowen Hou, Menglei Hua, Wei Zhang, Jianxin Ji, Xuan Zhang, Huiru Jiang, Mengyun Li, Xiaoxiao Wu, Wenwen Zhao, Shuxin Sun, Lei Cao, and Liuying Wang.
\newblock An ultrasonography of thyroid nodules dataset with pathological diagnosis annotation for deep learning.
\newblock \emph{Scientific Data}, 11\penalty0 (1):\penalty0 1272, 2024.

\bibitem[Hu et~al.(2024)Hu, Li, Lu, Shao, He, Qiao, and Luo]{hu2024omnimedvqa}
Yutao Hu, Tianbin Li, Quanfeng Lu, Wenqi Shao, Junjun He, Yu Qiao, and Ping Luo.
\newblock {OmniMedVQA}: A new large-scale comprehensive evaluation benchmark for medical {LVLM}.
\newblock In \emph{CVPR}, 2024.

\bibitem[Huang et~al.(2021)Huang, Yang, Liu, Tian, Liu, Wu, Lin, Wang, Morikawa, Chang, Tegner, and Worring]{huang2021deepeyenet}
Jia-Hong Huang, Chao-Han~Huck Yang, Fangyu Liu, Meng Tian, Yi-Chieh Liu, Ting-Wei Wu, I-Hung Lin, Kang Wang, Hiromasa Morikawa, Hernghua Chang, Jesper Tegner, and Marcel Worring.
\newblock {DeepOpht}: Medical report generation for retinal images via deep models and visual explanation.
\newblock In \emph{WACV}, 2021.

\bibitem[Huang et~al.(2024)Huang, Li, Lin, Zhang, Yi, and Wang]{huang2024hpmi}
Shenghai Huang, Zhongwen Li, Bing Lin, Shaodan Zhang, Quanyong Yi, and Lei Wang.
\newblock {HPMI}: A retinal fundus image dataset for identification of high and pathological myopia based on deep learning.
\newblock Figshare, \url{https://doi.org/10.6084/m9.figshare.24800232.v2}, 2024.

\bibitem[Ikezogwo et~al.(2023)Ikezogwo, Seyfioglu, Ghezloo, Geva, Mohammed, Anand, Krishna, and Shapiro]{ikezogwo2023quilt1m}
Wisdom Ikezogwo, Saygin Seyfioglu, Fatemeh Ghezloo, Dylan Geva, Fatwir~Sheikh Mohammed, Pavan~Kumar Anand, Ranjay Krishna, and Linda Shapiro.
\newblock {Quilt-1M}: One million image-text pairs for histopathology.
\newblock In \emph{NeurIPS}, 2023.

\bibitem[Iqbal and Sharif(2024)]{iqbal2024busuc}
Ahmed Iqbal and Muhammad Sharif.
\newblock Memory-efficient transformer network with feature fusion for breast tumor segmentation and classification task.
\newblock \emph{Engineering Applications of Artificial Intelligence}, 127:\penalty0 107292, 2024.

\bibitem[Jiang et~al.(2024)Jiang, Imran, Muralidharan, Patel, Pensa, Liang, Benidir, Grajo, Joseph, Terry, et~al.]{jiang2024mupsd}
Hongxu Jiang, Muhammad Imran, Preethika Muralidharan, Anjali Patel, Jake Pensa, Muxuan Liang, Tarik Benidir, Joseph~R. Grajo, Jason~P. Joseph, Russell Terry, et~al.
\newblock {MicroSegNet}: A deep learning approach for prostate segmentation on micro-ultrasound images.
\newblock \emph{Computerized Medical Imaging and Graphics}, 112:\penalty0 102326, 2024.

\bibitem[Jiang et~al.(2025)Jiang, Wang, Song, Hu, Zhou, Pu, Zhang, Yang, Feng, Zhou, Hao, Chen, Wu, Tang, Lv, Xu, Wang, Xiao, Feng, Zhu, Li, Xie, Sun, Wu, and Liu]{jiang2025hulumed}
Songtao Jiang, Yuan Wang, Sibo Song, Tianxiang Hu, Chenyi Zhou, Bin Pu, Yan Zhang, Zhibo Yang, Yang Feng, Joey~Tianyi Zhou, Jin Hao, Zijian Chen, Ruijia Wu, Tao Tang, Junhui Lv, Hongxia Xu, Hongwei Wang, Jun Xiao, Bin Feng, Fudong Zhu, Kenli Li, Weidi Xie, Jimeng Sun, Jian Wu, and Zuozhu Liu.
\newblock {Hulu-Med}: A transparent generalist model towards holistic medical vision-language understanding.
\newblock \emph{arXiv preprint arXiv:2510.08668}, 2025.

\bibitem[Karthik et~al.(2019)Karthik, Maggie, and Dane]{karthik2019aptos}
Karthik, Maggie, and Sohier Dane.
\newblock {APTOS} 2019 blindness detection.
\newblock Kaggle, \url{https://www.kaggle.com/competitions/aptos2019-blindness-detection}, 2019.

\bibitem[Kather et~al.(2019)Kather, Krisam, Charoentong, Luedde, Herpel, Weis, Gaiser, Marx, Valous, Ferber, et~al.]{kather2019crc100k}
Jakob~Nikolas Kather, Johannes Krisam, Pornpimol Charoentong, Tom Luedde, Esther Herpel, Cleo-Aron Weis, Timo Gaiser, Alexander Marx, Nektarios~A. Valous, Dyke Ferber, et~al.
\newblock Predicting survival from colorectal cancer histology slides using deep learning: A retrospective multicenter study.
\newblock \emph{PLOS Medicine}, 16\penalty0 (1):\penalty0 e1002730, 2019.

\bibitem[Katumba et~al.(2025)Katumba, Murindanyi, Okila, Nakatumba-Nabende, Mwikirize, Serugunda, Bugeza, Oriekot, Bossa, and Nabawanuka]{katumba2025lungus}
Andrew Katumba, Sudi Murindanyi, Nixson Okila, Joyce Nakatumba-Nabende, Cosmas Mwikirize, Jonathan Serugunda, Samuel Bugeza, Anthony Oriekot, Juliet Bossa, and Eva Nabawanuka.
\newblock A dataset of lung ultrasound images for automated {AI}-based lung disease classification.
\newblock \emph{Data in Brief}, 62:\penalty0 112034, 2025.

\bibitem[Kauppi et~al.(2006)Kauppi, Kalesnykiene, Kamarainen, Lensu, Sorri, Uusitalo, K{\"a}lvi{\"a}inen, and Pietil{\"a}]{kauppi2006diaretdb0}
Tomi Kauppi, Valentina Kalesnykiene, Joni-Kristian Kamarainen, Lasse Lensu, Iiris Sorri, Hannu Uusitalo, Heikki K{\"a}lvi{\"a}inen, and Juhani Pietil{\"a}.
\newblock {DIARETDB0}: Evaluation database and methodology for diabetic retinopathy algorithms.
\newblock Technical report, \url{https://www.it.lut.fi/project/imageret/diaretdb0/}, 2006.

\bibitem[Kermany et~al.(2018)Kermany, Goldbaum, Cai, Valentim, Liang, Baxter, McKeown, Yang, Wu, Yan, et~al.]{kermany2018oct2017}
Daniel~S. Kermany, Michael Goldbaum, Wenjia Cai, Carolina~C.S. Valentim, Huiying Liang, Sally~L. Baxter, Alex McKeown, Ge Yang, Xiaokang Wu, Fangbing Yan, et~al.
\newblock Identifying medical diagnoses and treatable diseases by image-based deep learning.
\newblock \emph{Cell}, 172\penalty0 (5):\penalty0 1122--1131.e9, 2018.

\bibitem[Kovalyk et~al.(2022)Kovalyk, Morales-S{\'a}nchez, Verd{\'u}-Monedero, Sell{\'e}s-Navarro, Palaz{\'o}n-Cabanes, and Sancho-G{\'o}mez]{kovalyk2022papila}
Oleksandr Kovalyk, Juan Morales-S{\'a}nchez, Rafael Verd{\'u}-Monedero, Inmaculada Sell{\'e}s-Navarro, Ana Palaz{\'o}n-Cabanes, and Jos{\'e}-Luis Sancho-G{\'o}mez.
\newblock {PAPILA}: Dataset with fundus images and clinical data of both eyes of the same patient for glaucoma assessment.
\newblock \emph{Scientific Data}, 9\penalty0 (1):\penalty0 291, 2022.

\bibitem[Kriegsmann et~al.(2022)Kriegsmann, Lobers, Zgorzelski, Kriegsmann, Jan{\ss}en, Meli{\ss}, Muley, Sack, Steinbuss, and Kriegsmann]{kriegsmann2022skincancer}
Katharina Kriegsmann, Frithjof Lobers, Christiane Zgorzelski, J{\"o}rg Kriegsmann, Charlotte Jan{\ss}en, Rolf~R{\"u}dinger Meli{\ss}, Thomas Muley, Ulrich Sack, Georg Steinbuss, and Mark Kriegsmann.
\newblock Deep learning for the detection of anatomical tissue structures and neoplasms of the skin on scanned histopathological tissue sections.
\newblock \emph{Frontiers in Oncology}, 12:\penalty0 1022967, 2022.

\bibitem[Kr{\"o}nke et~al.(2022)Kr{\"o}nke, Eilers, Dimova, K{\"o}hler, Buschner, Schweiger, Konstantinidou, Makowski, Nagarajah, Navab, Weber, and Wendler]{kronke2022segthy}
Markus Kr{\"o}nke, Christine Eilers, Desislava Dimova, Melanie K{\"o}hler, Gabriel Buschner, Lilit Schweiger, Lemonia Konstantinidou, Marcus Makowski, James Nagarajah, Nassir Navab, Wolfgang Weber, and Thomas Wendler.
\newblock Tracked {3D} ultrasound and deep neural network-based thyroid segmentation reduce interobserver variability in thyroid volumetry.
\newblock \emph{PLOS ONE}, 17\penalty0 (7):\penalty0 e0268550, 2022.

\bibitem[Kulyabin et~al.(2024)Kulyabin, Zhdanov, Nikiforova, Stepichev, Kuznetsova, Ronkin, Borisov, Bogachev, Korotkich, Constable, and Maier]{kulyabin2024octdl}
Mikhail Kulyabin, Aleksei Zhdanov, Anastasia Nikiforova, Andrey Stepichev, Anna Kuznetsova, Mikhail Ronkin, Vasilii Borisov, Alexander Bogachev, Sergey Korotkich, Paul~A. Constable, and Andreas Maier.
\newblock {OCTDL}: Optical coherence tomography dataset for image-based deep learning methods.
\newblock \emph{Scientific Data}, 11\penalty0 (1):\penalty0 365, 2024.

\bibitem[Kwon et~al.(2023)Kwon, Li, Zhuang, Sheng, Zheng, Yu, Gonzalez, Zhang, and Stoica]{kwon2023vllm}
Woosuk Kwon, Zhuohan Li, Siyuan Zhuang, Ying Sheng, Lianmin Zheng, Cody~Hao Yu, Joseph~E. Gonzalez, Hao Zhang, and Ion Stoica.
\newblock Efficient memory management for large language model serving with {PagedAttention}.
\newblock In \emph{SOSP}, 2023.

\bibitem[Laverde-Saad et~al.(2022)Laverde-Saad, Jfri, Garc{\'i}a, Salg{\"u}ero, Mart{\'i}nez, Cembrero, Roust{\'a}n, and Alfageme]{laverdesaad2022dermus}
Alexandra Laverde-Saad, Abdulhadi Jfri, Rub{\'e}n Garc{\'i}a, Irene Salg{\"u}ero, Constanza Mart{\'i}nez, Hirune Cembrero, Gast{\'o}n Roust{\'a}n, and Fernando Alfageme.
\newblock Discriminative deep learning based benignity/malignancy diagnosis of dermatologic ultrasound skin lesions with pretrained artificial intelligence architecture.
\newblock \emph{Skin Research and Technology}, 28\penalty0 (1):\penalty0 35--39, 2022.

\bibitem[Le et~al.(2026)Le, Liu, Wang, Liu, Zhu, Weng, Yu, Wang, Wu, Yan, Sun, Jiang, Pei, Liu, Zheng, Li, Noble, Souquet, Guo, Lin, and Guo]{le2026u2bench}
Anjie Le, Henan Liu, Yue Wang, Zhenyu Liu, Rongkun Zhu, Taohan Weng, Jinze Yu, Boyang Wang, Yalun Wu, Kaiwen Yan, Quanlin Sun, Meirui Jiang, Jialun Pei, Siya Liu, Haoyun Zheng, Zhoujun Li, J.~Alison Noble, Jacques Souquet, Xiaoqing Guo, Manxi Lin, and Hongcheng Guo.
\newblock {U2-BENCH}: Benchmarking large vision-language models on ultrasound understanding.
\newblock In \emph{ICLR}, 2026.

\bibitem[Leclerc et~al.(2019)Leclerc, Smistad, Pedrosa, Ostvik, Cervenansky, Espinosa, Espeland, Berg, Jodoin, Grenier, et~al.]{leclerc2019camus}
Sarah Leclerc, Erik Smistad, Joao Pedrosa, Andreas Ostvik, Frederic Cervenansky, Florian Espinosa, Torvald Espeland, Erik Andreas~Rye Berg, Pierre-Marc Jodoin, Thomas Grenier, et~al.
\newblock Deep learning for segmentation using an open large-scale dataset in {2D} echocardiography.
\newblock \emph{IEEE Transactions on Medical Imaging}, 38\penalty0 (9):\penalty0 2198--2210, 2019.

\bibitem[Li et~al.(2024)Li, Zhang, Wang, Zhu, Liu, Yang, Wang, Shen, and Sheng]{li2024lepset}
Jiajia Li, Pingping Zhang, Teng Wang, Lei Zhu, Ruhan Liu, Xia Yang, Kaixuan Wang, Dinggang Shen, and Bin Sheng.
\newblock {DSMT-Net}: Dual self-supervised multi-operator transformation for multi-source endoscopic ultrasound diagnosis.
\newblock \emph{IEEE Transactions on Medical Imaging}, 43\penalty0 (1):\penalty0 64--75, 2024.

\bibitem[Li et~al.(2021)Li, Li, Hu, Wang, and Kang]{li2021odir}
Ning Li, Tao Li, Chunyu Hu, Kai Wang, and Hong Kang.
\newblock A benchmark of ocular disease intelligent recognition: One shot for multi-disease detection.
\newblock In \emph{Bench}, 2021.

\bibitem[Li et~al.(2019)Li, Gao, Wang, Guo, Liu, and Kang]{li2019ddr}
Tao Li, Yingqi Gao, Kai Wang, Song Guo, Hanruo Liu, and Hong Kang.
\newblock Diagnostic assessment of deep learning algorithms for diabetic retinopathy screening.
\newblock \emph{Information Sciences}, 501:\penalty0 511--522, 2019.

\bibitem[Lin et~al.(2023)Lin, Zhao, Zhang, Wu, Zhang, Wang, and Xie]{lin2023pmcoa}
Weixiong Lin, Ziheng Zhao, Xiaoman Zhang, Chaoyi Wu, Ya Zhang, Yanfeng Wang, and Weidi Xie.
\newblock {PMC}-{CLIP}: Contrastive language-image pre-training using biomedical documents.
\newblock In \emph{MICCAI}, 2023.

\bibitem[Lin et~al.(2022)Lin, Lin, Zhu, Fu, Qin, and Wang]{lin2022cvanet}
Zhi Lin, Junhao Lin, Lei Zhu, Huazhu Fu, Jing Qin, and Liansheng Wang.
\newblock A new dataset and a baseline model for breast lesion detection in ultrasound videos.
\newblock In \emph{MICCAI}, 2022.

\bibitem[Liu et~al.(2022)Liu, Wang, Wu, Dai, Fang, Yan, Son, Tang, Li, Gao, et~al.]{liu2022deepdrid}
Ruhan Liu, Xiangning Wang, Qiang Wu, Ling Dai, Xi Fang, Tao Yan, Jaemin Son, Shiqi Tang, Jiang Li, Zijian Gao, et~al.
\newblock {DeepDRiD}: Diabetic retinopathy---grading and image quality estimation challenge.
\newblock \emph{Patterns}, 3\penalty0 (6):\penalty0 100512, 2022.

\bibitem[Lu et~al.(2025)Lu, She, He, Cheng, Wang, Huang, Lin, Lv, Qin, Liu, et~al.]{lu2025usliver}
Rui-Fang Lu, Chao-Yin She, Dan-Ni He, Mei-Qing Cheng, Ying Wang, Hui Huang, Ya-Dan Lin, Jia-Yi Lv, Si Qin, Ze-Zhi Liu, et~al.
\newblock {AI} enhanced diagnostic accuracy and workload reduction in hepatocellular carcinoma screening.
\newblock \emph{npj Digital Medicine}, 8\penalty0 (1):\penalty0 500, 2025.

\bibitem[Lu et~al.(2022)Lu, Zhou, Zhi, Zhou, Jiang, Qiu, Ou, Wang, Qiu, Zhong, et~al.]{lu2022jnuifm}
Yaosheng Lu, Mengqiang Zhou, Dengjiang Zhi, Minghong Zhou, Xiaosong Jiang, Ruiyu Qiu, Zhanhong Ou, Huijin Wang, Di Qiu, Mei Zhong, et~al.
\newblock The {JNU}-{IFM} dataset for segmenting pubic symphysis-fetal head.
\newblock \emph{Data in Brief}, 41:\penalty0 107904, 2022.

\bibitem[Marcinkevi{\v{c}}s et~al.(2024)Marcinkevi{\v{c}}s, Reis~Wolfertstetter, Klimiene, Chin-Cheong, Paschke, Zerres, Denzinger, Niederberger, Wellmann, Ozkan, Knorr, and Vogt]{marcinkevics2024appendicitis}
Ri{\v{c}}ards Marcinkevi{\v{c}}s, Patricia Reis~Wolfertstetter, Ugne Klimiene, Kieran Chin-Cheong, Alyssia Paschke, Julia Zerres, Markus Denzinger, David Niederberger, Sven Wellmann, Ece Ozkan, Christian Knorr, and Julia~E. Vogt.
\newblock Interpretable and intervenable ultrasonography-based machine learning models for pediatric appendicitis.
\newblock \emph{Medical Image Analysis}, 91:\penalty0 103042, 2024.

\bibitem[Mei et~al.(2022)Mei, Liu, Robson, Marinelli, Huang, Doshi, Jacobi, Cao, Link, Yang, et~al.]{mei2022radimagenet}
Xueyan Mei, Zelong Liu, Philip~M. Robson, Brett Marinelli, Mingqian Huang, Amish Doshi, Adam Jacobi, Chendi Cao, Katherine~E. Link, Thomas Yang, et~al.
\newblock {RadImageNet}: An open radiologic deep learning research dataset for effective transfer learning.
\newblock \emph{Radiology: Artificial Intelligence}, 4\penalty0 (5):\penalty0 e210315, 2022.

\bibitem[Meng et~al.(2025)Meng, Hao, Dai, Feng, Liu, Feng, Wu, Gai, Zhu, Hu, Wu, Xu, Li, Xiao, Liu, Zhou, Zhu, Zhao, Xia, Fang, Sun, Wu, and Liu]{meng2025dentvlm}
Zijie Meng, Jin Hao, Xiwei Dai, Yang Feng, Jiaxiang Liu, Bin Feng, Huikai Wu, Xiaotang Gai, Hengchuan Zhu, Tianxiang Hu, Yangyang Wu, Hongxia Xu, Jin Li, Jun Xiao, Xiaoqiang Liu, Joey~Tianyi Zhou, Fudong Zhu, Zhihe Zhao, Lunguo Xia, Bing Fang, Jimeng Sun, Jian Wu, and Zuozhu Liu.
\newblock {DentVLM}: A multimodal vision-language model for comprehensive dental diagnosis and enhanced clinical practice.
\newblock \emph{arXiv preprint arXiv:2509.23344}, 2025.

\bibitem[Mo et~al.(2023)Mo, Han, Liu, Liu, Shi, Lin, Zhao, Huang, Qiu, Cui, et~al.]{mo2023gdphsysucc}
Yuhao Mo, Chu Han, Yu Liu, Min Liu, Zhenwei Shi, Jiatai Lin, Bingchao Zhao, Chunwang Huang, Bingjiang Qiu, Yanfen Cui, et~al.
\newblock {HoVer-Trans}: Anatomy-aware {HoVer}-transformer for {ROI}-free breast cancer diagnosis in ultrasound images.
\newblock \emph{IEEE Transactions on Medical Imaging}, 42\penalty0 (6):\penalty0 1696--1706, 2023.

\bibitem[Naeem et~al.(2025)Naeem, Li, Liao, Xu, Mathew, Zhu, Tan, Jaiswal, Salibian, Hu, Chen, and Ding]{naeem2025pathrag}
Awais Naeem, Tianhao Li, Huang-Ru Liao, Jiawei Xu, Aby~Mammen Mathew, Zehao Zhu, Zhen Tan, Ajay~Kumar Jaiswal, Raffi~A. Salibian, Ziniu Hu, Tianlong Chen, and Ying Ding.
\newblock {Path-RAG}: Knowledge-guided key region retrieval for open-ended pathology visual question answering.
\newblock In \emph{ML4H}, 2025.

\bibitem[Nelson(2023)]{nelson2023knee}
Amanda Nelson.
\newblock Development of an {AI/ML}-ready knee ultrasound dataset in a population-based cohort.
\newblock Harvard Dataverse, \url{https://doi.org/10.7910/DVN/SKP9IB}, 2023.

\bibitem[Pachade et~al.(2021)Pachade, Porwal, Thulkar, Kokare, Deshmukh, Sahasrabuddhe, Giancardo, Quellec, and M{\'e}riaudeau]{pachade2021rfmid}
Samiksha Pachade, Prasanna Porwal, Dhanshree Thulkar, Manesh Kokare, Girish Deshmukh, Vivek Sahasrabuddhe, Luca Giancardo, Gwenol{\'e} Quellec, and Fabrice M{\'e}riaudeau.
\newblock Retinal fundus multi-disease image dataset ({RFMiD}): A dataset for multi-disease detection research.
\newblock \emph{Data}, 6\penalty0 (2):\penalty0 14, 2021.

\bibitem[{Pahuni Choudhary}(2024)]{choudhary2024carotid}
{Pahuni Choudhary}.
\newblock Carotid artery ultrasound and color {Doppler}.
\newblock Kaggle, \url{https://www.kaggle.com/datasets/pahunichoudhary/carotid-artery-ultrasound-and-color-doppler}, 2024.

\bibitem[Paw{\l}owska et~al.(2024)Paw{\l}owska, {\'C}wierz-Pie{\'n}kowska, Domalik, Jagu{\'s}, Kasprzak, Matkowski, Fura, Nowicki, and {\.Z}o{\l}ek]{pawlowska2024breast}
Anna Paw{\l}owska, Anna {\'C}wierz-Pie{\'n}kowska, Agnieszka Domalik, Dominika Jagu{\'s}, Piotr Kasprzak, Rafa{\l} Matkowski, {\L}ukasz Fura, Andrzej Nowicki, and Norbert {\.Z}o{\l}ek.
\newblock Curated benchmark dataset for ultrasound based breast lesion analysis.
\newblock \emph{Scientific Data}, 11\penalty0 (1):\penalty0 148, 2024.

\bibitem[Porwal et~al.(2018)Porwal, Pachade, Kamble, Kokare, Deshmukh, Sahasrabuddhe, and Meriaudeau]{porwal2018idrid}
Prasanna Porwal, Samiksha Pachade, Ravi Kamble, Manesh Kokare, Girish Deshmukh, Vivek Sahasrabuddhe, and Fabrice Meriaudeau.
\newblock {Indian} diabetic retinopathy image dataset ({IDRiD}): A database for diabetic retinopathy screening research.
\newblock \emph{Data}, 3\penalty0 (3):\penalty0 25, 2018.

\bibitem[Prabakaran et~al.(2023)Prabakaran, Hamelmann, Ostrowski, and Shafique]{prabakaran2023fpus23}
Bharath~Srinivas Prabakaran, Paul Hamelmann, Erik Ostrowski, and Muhammad Shafique.
\newblock {FPUS23}: An ultrasound fetus phantom dataset with deep neural network evaluations for fetus orientations, fetal planes, and anatomical features.
\newblock \emph{IEEE Access}, 11:\penalty0 58308--58317, 2023.

\bibitem[{Qwen Team}(2026)]{qwen35}
{Qwen Team}.
\newblock {Qwen3.5}: Towards native multimodal agents.
\newblock \url{https://qwen.ai/blog?id=qwen3.5}, 2026.

\bibitem[R{\"u}ckert et~al.(2024)R{\"u}ckert, Bloch, Br{\"u}ngel, Idrissi-Yaghir, Sch{\"a}fer, Schmidt, Koitka, Pelka, Abacha, G.~Seco~de Herrera, et~al.]{ruckert2024rocov2}
Johannes R{\"u}ckert, Louise Bloch, Raphael Br{\"u}ngel, Ahmad Idrissi-Yaghir, Henning Sch{\"a}fer, Cynthia~S. Schmidt, Sven Koitka, Obioma Pelka, Asma~Ben Abacha, Alba G.~Seco~de Herrera, et~al.
\newblock {ROCOv2}: Radiology objects in {COntext} version 2, an updated multimodal image dataset.
\newblock \emph{Scientific Data}, 11\penalty0 (1):\penalty0 688, 2024.

\bibitem[Seyfioglu et~al.(2024)Seyfioglu, Ikezogwo, Ghezloo, Krishna, and Shapiro]{seyfioglu2024quiltllava}
Mehmet~Saygin Seyfioglu, Wisdom~O. Ikezogwo, Fatemeh Ghezloo, Ranjay Krishna, and Linda Shapiro.
\newblock {Quilt-LLaVA}: Visual instruction tuning by extracting localized narratives from open-source histopathology videos.
\newblock In \emph{CVPR}, 2024.

\bibitem[She et~al.(2026)She, Lu, Chen, Wang, and Huang]{she2026echovlm}
Chaoyin She, Ruifang Lu, Lida Chen, Wei Wang, and Qinghua Huang.
\newblock {EchoVLM}: Dynamic mixture-of-experts vision-language model for universal ultrasound intelligence.
\newblock In \emph{ACL}, 2026.

\bibitem[Silva-Rodr{\'i}guez et~al.(2020)Silva-Rodr{\'i}guez, Colomer, Sales, Molina, and Naranjo]{silvarodriguez2020sicapv2}
Julio Silva-Rodr{\'i}guez, Adri{\'a}n Colomer, Mar{\'i}a~A. Sales, Rafael Molina, and Valery Naranjo.
\newblock Going deeper through the {Gleason} scoring scale: An automatic end-to-end system for histology prostate grading and cribriform pattern detection.
\newblock \emph{Computer Methods and Programs in Biomedicine}, 195:\penalty0 105637, 2020.

\bibitem[Sotoudeh-Paima et~al.(2022)Sotoudeh-Paima, Jodeiri, Hajizadeh, and Soltanian-Zadeh]{sotoudehpaima2022neh}
Saman Sotoudeh-Paima, Ata Jodeiri, Fedra Hajizadeh, and Hamid Soltanian-Zadeh.
\newblock Multi-scale convolutional neural network for automated {AMD} classification using retinal {OCT} images.
\newblock \emph{Computers in Biology and Medicine}, 144:\penalty0 105368, 2022.

\bibitem[Spanhol et~al.(2016)Spanhol, Oliveira, Petitjean, and Heutte]{spanhol2016breakhis}
Fabio~A. Spanhol, Luiz~S. Oliveira, Caroline Petitjean, and Laurent Heutte.
\newblock A dataset for breast cancer histopathological image classification.
\newblock \emph{IEEE Transactions on Biomedical Engineering}, 63\penalty0 (7):\penalty0 1455--1462, 2016.

\bibitem[Sun et~al.(2024)Sun, Wu, Zhu, Zheng, Chen, Zhang, Zhang, Wan, Lan, Zheng, Li, Lyu, Lin, and Yang]{sun2024pathmmu}
Yuxuan Sun, Hao Wu, Chenglu Zhu, Sunyi Zheng, Qizi Chen, Kai Zhang, Yunlong Zhang, Dan Wan, Xiaoxiao Lan, Mengyue Zheng, Jingxiong Li, Xinheng Lyu, Tao Lin, and Lin Yang.
\newblock {PathMMU}: A massive multimodal expert-level benchmark for understanding and reasoning in pathology.
\newblock In \emph{ECCV}, 2024.

\bibitem[Tak et~al.(2026)Tak, Ko, Kwon, Abbas, Cho, Kim, Seo, Oh, and Lee]{tak2026smclud}
Jisoo Tak, Ryoung-Eun Ko, Ryan~Donghan Kwon, Zeeshan Abbas, Yang~Hyun Cho, Jongman Kim, Seondeok Seo, Namkee Oh, and Seung~Won Lee.
\newblock {SMC}-{LUD}: Large-scale {B}-mode liver ultrasound dataset for hepatocellular carcinoma and hemangioma classification.
\newblock \emph{Scientific Data}, 13\penalty0 (1):\penalty0 649, 2026.

\bibitem[Tascon-Morales et~al.(2022)Tascon-Morales, M{\'a}rquez-Neila, and Sznitman]{tasconmorales2022dmevqa}
Sergio Tascon-Morales, Pablo M{\'a}rquez-Neila, and Raphael Sznitman.
\newblock Consistency-preserving visual question answering in medical imaging.
\newblock In \emph{MICCAI}, 2022.

\bibitem[Tolkach et~al.(2023)Tolkach, Wolgast, Damanakis, Pryalukhin, Schallenberg, Hulla, Eich, Schroeder, Mukhopadhyay, Fuchs, et~al.]{tolkach2023esca}
Yuri Tolkach, Lisa~Marie Wolgast, Alexander Damanakis, Alexey Pryalukhin, Simon Schallenberg, Wolfgang Hulla, Marie-Lisa Eich, Wolfgang Schroeder, Anirban Mukhopadhyay, Moritz Fuchs, et~al.
\newblock Artificial intelligence for tumour tissue detection and histological regression grading in oesophageal adenocarcinomas: a retrospective algorithm development and validation study.
\newblock \emph{The Lancet Digital Health}, 5\penalty0 (5):\penalty0 e265--e275, 2023.

\bibitem[Turki et~al.(2024)Turki, Obaid, Bellaaj, Ksantini, and AlTaee]{turki2024uidatagb}
Amina Turki, Ahmed~Mahdi Obaid, Hatem Bellaaj, Mohamed Ksantini, and Abdulla AlTaee.
\newblock {UIdataGB}: Multi-class ultrasound images dataset for gallbladder disease detection.
\newblock \emph{Data in Brief}, 54:\penalty0 110426, 2024.

\bibitem[Vallez et~al.(2025)Vallez, Bueno, Deniz, Rienda, and Pastor]{vallez2025busuclm}
Noelia Vallez, Gloria Bueno, Oscar Deniz, Miguel~Angel Rienda, and Carlos Pastor.
\newblock {BUS}-{UCLM}: Breast ultrasound lesion segmentation dataset.
\newblock \emph{Scientific Data}, 12\penalty0 (1):\penalty0 242, 2025.

\bibitem[Veeling et~al.(2018)Veeling, Linmans, Winkens, Cohen, and Welling]{veeling2018pcam}
Bastiaan~S. Veeling, Jasper Linmans, Jim Winkens, Taco Cohen, and Max Welling.
\newblock Rotation equivariant {CNNs} for digital pathology.
\newblock In \emph{MICCAI}, 2018.

\bibitem[{Vuppala Adithya Sairam}(2022)]{sairam2022ubibc}
{Vuppala Adithya Sairam}.
\newblock Ultrasound breast images for breast cancer.
\newblock Kaggle, \url{https://www.kaggle.com/datasets/vuppalaadithyasairam/ultrasound-breast-images-for-breast-cancer}, 2022.

\bibitem[Wang et~al.(2023)Wang, Wang, Wang, Li, Da, Liu, Gao, Shen, He, Shen, Duan, Zhao, Li, Qiao, and Zhang]{wang2023medfmc}
Dequan Wang, Xiaosong Wang, Lilong Wang, Mengzhang Li, Qian Da, Xiaoqiang Liu, Xiangyu Gao, Jun Shen, Junjun He, Tian Shen, Qi Duan, Jie Zhao, Kang Li, Yu Qiao, and Shaoting Zhang.
\newblock {MedFMC}: A real-world dataset and benchmark for foundation model adaptation in medical image classification.
\newblock \emph{Scientific Data}, 10:\penalty0 574, 2023.

\bibitem[Wang et~al.(2025)Wang, Qi, Ou, An, Jin, Kong, and Li]{wang2025multieye}
Lehan Wang, Chongchong Qi, Chubin Ou, Lin An, Mei Jin, Xiangbin Kong, and Xiaomeng Li.
\newblock {MultiEYE}: Dataset and benchmark for {OCT}-enhanced retinal disease recognition from fundus images.
\newblock \emph{IEEE Transactions on Medical Imaging}, 44\penalty0 (4):\penalty0 1711--1722, 2025.

\bibitem[Wang et~al.(2026)Wang, Wu, Zhang, Lai, Lou, and Kong]{wang2026sonoinstruct}
Shijie Wang, Yingnan Wu, Yilun Zhang, Zeyu Lai, Wei Lou, and Dexing Kong.
\newblock A multimodal instruction dataset and benchmark for ultrasound understanding.
\newblock \emph{npj Digital Medicine}, 9\penalty0 (1):\penalty0 799, 2026.

\bibitem[Wang et~al.(2022)Wang, Li, Cui, Zhang, Luo, Yang, and Chang]{wang2022kfgnet}
Yuchen Wang, Zhongyu Li, Xiangxiang Cui, Liangliang Zhang, Xiang Luo, Meng Yang, and Shi Chang.
\newblock Key-frame guided network for thyroid nodule recognition using ultrasound videos.
\newblock In \emph{MICCAI}, 2022.

\bibitem[Wei et~al.(2021{\natexlab{a}})Wei, Suriawinata, Ren, Liu, Lisovsky, Vaickus, Brown, Baker, Tomita, Torresani, Wei, and Hassanpour]{wei2021mhist}
Jerry Wei, Arief Suriawinata, Bing Ren, Xiaoying Liu, Mikhail Lisovsky, Louis Vaickus, Charles Brown, Michael Baker, Naofumi Tomita, Lorenzo Torresani, Jason Wei, and Saeed Hassanpour.
\newblock A {Petri} dish for histopathology image analysis.
\newblock In \emph{AIME}, 2021{\natexlab{a}}.

\bibitem[Wei et~al.(2021{\natexlab{b}})Wei, Li, Yu, Zhang, Zhang, Hu, Mo, Gong, Chen, Ding, and Chen]{wei2021retinallesions}
Qijie Wei, Xirong Li, Weihong Yu, Xiao Zhang, Yongpeng Zhang, Bojie Hu, Bin Mo, Di Gong, Ning Chen, Dayong Ding, and Youxin Chen.
\newblock Learn to segment retinal lesions and beyond.
\newblock In \emph{ICPR}, 2021{\natexlab{b}}.

\bibitem[Wei et~al.(2025)Wei, Qian, and Li]{wei2025funbench}
Qijie Wei, Kaiheng Qian, and Xirong Li.
\newblock {FunBench}: Benchmarking fundus reading skills of {MLLMs}.
\newblock In \emph{MICCAI}, 2025.

\bibitem[Wiedemann et~al.(2025)Wiedemann, de~Korte-de Boer, Richter, van~de Weijer, Buhre, Eggert, Aarnoudse, Grevendonk, R{\"o}ber, Remie, et~al.]{wiedemann2025covidblues}
Nina Wiedemann, Dianne de~Korte-de Boer, Matthias Richter, Sjors van~de Weijer, Charlotte Buhre, Franz A.~M. Eggert, Sophie Aarnoudse, Lotte Grevendonk, Steffen R{\"o}ber, Carlijn~M.E. Remie, et~al.
\newblock {COVID}-{BLUeS} - a prospective study on the value of {AI} in lung ultrasound analysis.
\newblock \emph{IEEE Journal of Biomedical and Health Informatics}, 29\penalty0 (9):\penalty0 6301--6310, 2025.

\bibitem[Wu et~al.(2024)Wu, Zhang, Zhang, Zhou, Zhou, and Fu]{wu2024mmretinal}
Ruiqi Wu, Chenran Zhang, Jianle Zhang, Yi Zhou, Tao Zhou, and Huazhu Fu.
\newblock {MM-Retinal}: Knowledge-enhanced foundational pretraining with fundus image-text expertise.
\newblock In \emph{MICCAI}, 2024.

\bibitem[Xu et~al.(2023)Xu, Zheng, Liu, Wu, Ju, Wang, Lian, Zhang, Liang, Sang, et~al.]{xu2023aul}
Yiming Xu, Bowen Zheng, Xiaohong Liu, Tao Wu, Jinxiu Ju, Shijie Wang, Yufan Lian, Hongjun Zhang, Tong Liang, Ye Sang, et~al.
\newblock Improving artificial intelligence pipeline for liver malignancy diagnosis using ultrasound images and video frames.
\newblock \emph{Briefings in Bioinformatics}, 24\penalty0 (1):\penalty0 bbac569, 2023.

\bibitem[Yu et~al.(2026)Yu, Li, Niu, Zhang, Gong, Li, Zou, Qi, Cao, Lan, et~al.]{yu2026buscot}
Haojun Yu, Youcheng Li, Zihan Niu, Nan Zhang, Xuantong Gong, Huan Li, Zhiying Zou, Haifeng Qi, Zhenxiao Cao, Zijie Lan, et~al.
\newblock A chain-of-thought reasoning breast ultrasound dataset covering all histopathology categories.
\newblock \emph{Scientific Data}, 13\penalty0 (1):\penalty0 370, 2026.

\bibitem[Zhang et~al.(2025)Zhang, Liu, Han, Niu, and Sun]{zhang2025tn5000}
Huan Zhang, Qianglin Liu, Xiaolin Han, Lijuan Niu, and Weidong Sun.
\newblock {TN5000}: An ultrasound image dataset for thyroid nodule detection and classification.
\newblock \emph{Scientific Data}, 12\penalty0 (1):\penalty0 1437, 2025.

\bibitem[Zhang et~al.(2024)Zhang, Wu, Zhao, Lin, Zhang, Wang, and Xie]{zhang2024pmcvqa}
Xiaoman Zhang, Chaoyi Wu, Ziheng Zhao, Weixiong Lin, Ya Zhang, Yanfeng Wang, and Weidi Xie.
\newblock Development of a large-scale medical visual question-answering dataset.
\newblock \emph{Communications Medicine}, 4\penalty0 (1):\penalty0 277, 2024.

\bibitem[Zhao et~al.(2022)Zhao, Lyu, Bai, Cai, Liu, Cheng, Wu, Sang, Yang, and Chen]{zhao2022mmotu}
Qi Zhao, Shuchang Lyu, Wenpei Bai, Linghan Cai, Binghao Liu, Guangliang Cheng, Meijing Wu, Xiubo Sang, Min Yang, and Lijiang Chen.
\newblock {MMOTU}: A multi-modality ovarian tumor ultrasound image dataset for unsupervised cross-domain semantic segmentation.
\newblock \emph{arXiv preprint arXiv:2207.06799}, 2022.

\bibitem[Zhu et~al.(2022)Zhu, Chen, Peng, Wang, and Jin]{zhu2022chaoyang}
Chuang Zhu, Wenkai Chen, Ting Peng, Ying Wang, and Mulan Jin.
\newblock Hard sample aware noise robust learning for histopathology image classification.
\newblock \emph{IEEE Transactions on Medical Imaging}, 41\penalty0 (4):\penalty0 881--894, 2022.

\end{thebibliography}


\begin{thebibliography}{46}
\providecommand{\natexlab}[1]{#1}
\providecommand{\url}[1]{\texttt{#1}}
\expandafter\ifx\csname urlstyle\endcsname\relax
  \providecommand{\doi}[1]{doi: #1}\else
  \providecommand{\doi}{doi: \begingroup \urlstyle{rm}\Url}\fi

\bibitem[Agarwal et~al.(2024)Agarwal, Vieillard, Zhou, Stanczyk, Ramos, Geist, and Bachem]{agarwal2024gkd}
Rishabh Agarwal, Nino Vieillard, Yongchao Zhou, Piotr Stanczyk, Sabela Ramos, Matthieu Geist, and Olivier Bachem.
\newblock On-policy distillation of language models: Learning from self-generated mistakes.
\newblock In \emph{ICLR}, 2024.

\bibitem[Bai et~al.(2025)Bai, Cai, Chen, Chen, Chen, Cheng, Deng, Ding, Gao, Ge, et~al.]{bai2025qwen3vl}
Shuai Bai, Yuxuan Cai, Ruizhe Chen, Keqin Chen, Xionghui Chen, Zesen Cheng, Lianghao Deng, Wei Ding, Chang Gao, Chunjiang Ge, et~al.
\newblock {Qwen3-VL} technical report.
\newblock \emph{arXiv preprint arXiv:2511.21631}, 2025.

\bibitem[Bengio et~al.(2015)Bengio, Vinyals, Jaitly, and Shazeer]{bengio2015scheduled}
Samy Bengio, Oriol Vinyals, Navdeep Jaitly, and Noam Shazeer.
\newblock Scheduled sampling for sequence prediction with recurrent neural networks.
\newblock In \emph{NeurIPS}, 2015.

\bibitem[Boizard et~al.(2025)Boizard, El~Haddad, Hudelot, and Colombo]{boizard2024uld}
Nicolas Boizard, Kevin El~Haddad, C{\'e}line Hudelot, and Pierre Colombo.
\newblock Towards cross-tokenizer distillation: the universal logit distillation loss for {LLMs}.
\newblock \emph{Transactions on Machine Learning Research}, 2025.

\bibitem[Bousselham et~al.(2026)Bousselham, Kuehne, and Schmid]{bousselham2026vold}
Walid Bousselham, Hilde Kuehne, and Cordelia Schmid.
\newblock {VOLD}: Reasoning transfer from {LLMs} to vision-language models via on-policy distillation.
\newblock In \emph{CVPR}, 2026.

\bibitem[Deng et~al.(2026)Deng, Wei, Qian, Liu, Lan, Shen, Dong, Yu, and Li]{deng2026fundusr1}
Yuchuan Deng, Qijie Wei, Kaiheng Qian, Jiazhen Liu, Bangxiang Lan, Qingwei Shen, Jianfeng Dong, Weihong Yu, and Xirong Li.
\newblock {Fundus-R1}: Training a fundus-reading {MLLM} on public data.
\newblock \emph{arXiv preprint arXiv:2604.08322}, 2026.

\bibitem[Gu et~al.(2024)Gu, Dong, Wei, and Huang]{gu2024minillm}
Yuxian Gu, Li Dong, Furu Wei, and Minlie Huang.
\newblock {MiniLLM}: Knowledge distillation of large language models.
\newblock In \emph{ICLR}, 2024.

\bibitem[Guo et~al.(2024)Guo, Chai, Li, and Wang]{guo2024llavaultra}
Xuechen Guo, Wenhao Chai, Shi-Yan Li, and Gaoang Wang.
\newblock {LLaVA-Ultra}: Large {Chinese} language and vision assistant for ultrasound.
\newblock In \emph{ACM MM}, 2024.

\bibitem[Hao et~al.(2025)Hao, Liang, Lin, Fan, Zhou, Guo, Ye, Sun, Zhang, Yang, Li, Tang, Tsoi, Shen, and Hung]{hao2025oralgptomni}
Jing Hao, Yuci Liang, Lizhuo Lin, Yuxuan Fan, Wenkai Zhou, Kaixin Guo, Zanting Ye, Yanpeng Sun, Xinyu Zhang, Yanqi Yang, Qiankun Li, Hao Tang, James Kit-Hon Tsoi, Linlin Shen, and Kuo~Feng Hung.
\newblock {OralGPT-Omni}: A versatile dental multimodal large language model.
\newblock \emph{arXiv preprint arXiv:2511.22055}, 2025.

\bibitem[He et~al.(2020)He, Zhang, Mou, Xing, and Xie]{he2020pathvqa}
Xuehai He, Yichen Zhang, Luntian Mou, Eric Xing, and Pengtao Xie.
\newblock {PathVQA}: 30000+ questions for medical visual question answering.
\newblock \emph{arXiv preprint arXiv:2003.10286}, 2020.

\bibitem[Hu et~al.(2024)Hu, Li, Lu, Shao, He, Qiao, and Luo]{hu2024omnimedvqa}
Yutao Hu, Tianbin Li, Quanfeng Lu, Wenqi Shao, Junjun He, Yu Qiao, and Ping Luo.
\newblock {OmniMedVQA}: A new large-scale comprehensive evaluation benchmark for medical {LVLM}.
\newblock In \emph{CVPR}, 2024.

\bibitem[Jiang et~al.(2025)Jiang, Wang, Song, Hu, Zhou, Pu, Zhang, Yang, Feng, Zhou, Hao, Chen, Wu, Tang, Lv, Xu, Wang, Xiao, Feng, Zhu, Li, Xie, Sun, Wu, and Liu]{jiang2025hulumed}
Songtao Jiang, Yuan Wang, Sibo Song, Tianxiang Hu, Chenyi Zhou, Bin Pu, Yan Zhang, Zhibo Yang, Yang Feng, Joey~Tianyi Zhou, Jin Hao, Zijian Chen, Ruijia Wu, Tao Tang, Junhui Lv, Hongxia Xu, Hongwei Wang, Jun Xiao, Bin Feng, Fudong Zhu, Kenli Li, Weidi Xie, Jimeng Sun, Jian Wu, and Zuozhu Liu.
\newblock {Hulu-Med}: A transparent generalist model towards holistic medical vision-language understanding.
\newblock \emph{arXiv preprint arXiv:2510.08668}, 2025.

\bibitem[Kim and Rush(2016)]{kim2016sequence}
Yoon Kim and Alexander~M. Rush.
\newblock Sequence-level knowledge distillation.
\newblock In \emph{EMNLP}, 2016.

\bibitem[Le et~al.(2026)Le, Liu, Wang, Liu, Zhu, Weng, Yu, Wang, Wu, Yan, Sun, Jiang, Pei, Liu, Zheng, Li, Noble, Souquet, Guo, Lin, and Guo]{le2026u2bench}
Anjie Le, Henan Liu, Yue Wang, Zhenyu Liu, Rongkun Zhu, Taohan Weng, Jinze Yu, Boyang Wang, Yalun Wu, Kaiwen Yan, Quanlin Sun, Meirui Jiang, Jialun Pei, Siya Liu, Haoyun Zheng, Zhoujun Li, J.~Alison Noble, Jacques Souquet, Xiaoqing Guo, Manxi Lin, and Hongcheng Guo.
\newblock {U2-BENCH}: Benchmarking large vision-language models on ultrasound understanding.
\newblock In \emph{ICLR}, 2026.

\bibitem[Li et~al.(2023)Li, Wong, Zhang, Usuyama, Liu, Yang, Naumann, Poon, and Gao]{li2023llavamed}
Chunyuan Li, Cliff Wong, Sheng Zhang, Naoto Usuyama, Haotian Liu, Jianwei Yang, Tristan Naumann, Hoifung Poon, and Jianfeng Gao.
\newblock {LLaVA-Med}: Training a large language-and-vision assistant for biomedicine in one day.
\newblock In \emph{NeurIPS}, 2023.

\bibitem[Lin et~al.(2025)Lin, Zhang, Li, Yuan, Yu, Li, He, Jiang, Li, Song, Tang, Xiao, Lin, Zhuang, and Ooi]{lin2025healthgpt}
Tianwei Lin, Wenqiao Zhang, Sijing Li, Yuqian Yuan, Binhe Yu, Haoyuan Li, Wanggui He, Hao Jiang, Mengze Li, Xiaohui Song, Siliang Tang, Jun Xiao, Hui Lin, Yueting Zhuang, and Beng~Chin Ooi.
\newblock {HealthGPT}: A medical large vision-language model for unifying comprehension and generation via heterogeneous knowledge adaptation.
\newblock In \emph{ICML}, 2025.

\bibitem[Lu and {Thinking Machines Lab}(2025)]{lu2025onpolicy}
Kevin Lu and {Thinking Machines Lab}.
\newblock On-policy distillation.
\newblock Thinking Machines Lab: Connectionism, \url{https://thinkingmachines.ai/blog/on-policy-distillation/}, 2025.

\bibitem[Lu et~al.(2024)Lu, Chen, Williamson, Chen, Zhao, Chow, Ikemura, Kim, Pouli, Patel, Soliman, Chen, Ding, Wang, Gerber, Liang, Le, Parwani, Weishaupt, and Mahmood]{lu2024pathchat}
Ming~Y. Lu, Bowen Chen, Drew F.~K. Williamson, Richard~J. Chen, Melissa Zhao, Aaron~K. Chow, Kenji Ikemura, Ahrong Kim, Dimitra Pouli, Ankush Patel, Amr Soliman, Chengkuan Chen, Tong Ding, Judy~J. Wang, Georg Gerber, Ivy Liang, Long~Phi Le, Anil~V. Parwani, Luca~L. Weishaupt, and Faisal Mahmood.
\newblock A multimodal generative {AI} copilot for human pathology.
\newblock \emph{Nature}, 634\penalty0 (8033):\penalty0 466--473, 2024.

\bibitem[Ma et~al.(2026)Ma, Wei, Zhao, Zhang, Xiao, Li, Yang, Gao, Wang, Li, Dong, Sui, and Luo]{ma2026mopd}
Wenhan Ma, Jianyu Wei, Liang Zhao, Hailin Zhang, Bangjun Xiao, Lei Li, Qibin Yang, Bofei Gao, Yudong Wang, Rang Li, Jinhao Dong, Zhifang Sui, and Fuli Luo.
\newblock {MOPD}: Multi-teacher on-policy distillation for capability integration in {LLM} post-training.
\newblock \emph{arXiv preprint arXiv:2606.30406}, 2026.

\bibitem[Meng et~al.(2025)Meng, Hao, Dai, Feng, Liu, Feng, Wu, Gai, Zhu, Hu, Wu, Xu, Li, Xiao, Liu, Zhou, Zhu, Zhao, Xia, Fang, Sun, Wu, and Liu]{meng2025dentvlm}
Zijie Meng, Jin Hao, Xiwei Dai, Yang Feng, Jiaxiang Liu, Bin Feng, Huikai Wu, Xiaotang Gai, Hengchuan Zhu, Tianxiang Hu, Yangyang Wu, Hongxia Xu, Jin Li, Jun Xiao, Xiaoqiang Liu, Joey~Tianyi Zhou, Fudong Zhu, Zhihe Zhao, Lunguo Xia, Bing Fang, Jimeng Sun, Jian Wu, and Zuozhu Liu.
\newblock {DentVLM}: A multimodal vision-language model for comprehensive dental diagnosis and enhanced clinical practice.
\newblock \emph{arXiv preprint arXiv:2509.23344}, 2025.

\bibitem[Minixhofer et~al.(2025)Minixhofer, Vuli{\'c}, and Ponti]{minixhofer2025alm}
Benjamin Minixhofer, Ivan Vuli{\'c}, and Edoardo~Maria Ponti.
\newblock Universal cross-tokenizer distillation via approximate likelihood matching.
\newblock In \emph{NeurIPS}, 2025.

\bibitem[Mou et~al.(2026)Mou, Zhang, Li, Yan, Hu, Shi, Xiong, Zhu, Li, and Chen]{mou2026mergemedbench}
Lichao Mou, Shilan Zhang, Chunlei Li, Bingcong Yan, Jingliang Hu, Yilei Shi, Shengwu Xiong, Xiao~Xiang Zhu, Lei Li, and Yaxiong Chen.
\newblock Model merging for medical {LVLMs}: A benchmark and a winner-take-all approach.
\newblock \emph{arXiv preprint arXiv:2607.15661}, 2026.

\bibitem[Nath et~al.(2025)Nath, Li, Yang, Myronenko, Zheng, Lu, Liu, Yin, Law, Tang, et~al.]{nath2025vilam3}
Vishwesh Nath, Wenqi Li, Dong Yang, Andriy Myronenko, Mingxin Zheng, Yao Lu, Zhijian Liu, Hongxu Yin, Yee~Man Law, Yucheng Tang, et~al.
\newblock {VILA-M3}: Enhancing vision-language models with medical expert knowledge.
\newblock In \emph{CVPR}, 2025.

\bibitem[Niu et~al.(2026)Niu, Xiao, Liu, Wang, Gong, Wang, and Li]{niu2026breaking}
Yifan Niu, Han Xiao, Dongyi Liu, Zelong Wang, Dihong Gong, Yasheng Wang, and Jia Li.
\newblock Breaking the tokenizer barrier: On-policy distillation across model families.
\newblock \emph{arXiv preprint arXiv:2606.09456}, 2026.

\bibitem[Pati{\~n}o et~al.(2025)Pati{\~n}o, Rasul, Gallou{\'e}dec, Burtenshaw, Paniego, Srivastav, Frere, Beeching, Tunstall, von Werra, and Wolf]{gold2025}
Carlos~Miguel Pati{\~n}o, Kashif Rasul, Quentin Gallou{\'e}dec, Ben Burtenshaw, Sergio Paniego, Vaibhav Srivastav, Thibaud Frere, Ed Beeching, Lewis Tunstall, Leandro von Werra, and Thomas Wolf.
\newblock Unlocking on-policy distillation for any model family.
\newblock Hugging Face Blog, \url{https://huggingface.co/spaces/HuggingFaceH4/on-policy-distillation}, 2025.

\bibitem[Qian et~al.(2026)Qian, Yu, Liu, Wang, Li, and Hu]{qian2026medopd}
Yunhang Qian, Jiaquan Yu, Jiawei Liu, Meng Wang, Hongwei~Bran Li, and Xiaobin Hu.
\newblock {Med-OPD}: Improving medical vision-language models via evidence-aware on-policy distillation.
\newblock \emph{arXiv preprint arXiv:2607.16303}, 2026.

\bibitem[Qin et~al.(2026)Qin, Shi, Han, Qian, Jie, and Ma]{qin2026videomopd}
Zhenxin Qin, Peng Shi, Cong Han, Yinlong Qian, Zequn Jie, and Lin Ma.
\newblock {Video-MOPD}: Multi-teacher on-policy distillation for video understanding.
\newblock \emph{arXiv preprint arXiv:2609.09300}, 2026.

\bibitem[{Qwen Team}(2026{\natexlab{a}})]{qwen35}
{Qwen Team}.
\newblock {Qwen3.5}: Towards native multimodal agents.
\newblock \url{https://qwen.ai/blog?id=qwen3.5}, 2026{\natexlab{a}}.

\bibitem[{Qwen Team}(2026{\natexlab{b}})]{qwen38}
{Qwen Team}.
\newblock {Qwen3.8-27B}.
\newblock \url{https://huggingface.co/Qwen/Qwen3.8-27B}, 2026{\natexlab{b}}.
\newblock Model card.

\bibitem[Sellergren et~al.(2026{\natexlab{a}})Sellergren, Gao, Mahvar, Kohlberger, Jamil, Traverse, Tono, Sadjad, Yang, Lau, et~al.]{sellergren2026medgemma15}
Andrew Sellergren, Chufan Gao, Fereshteh Mahvar, Timo Kohlberger, Fayaz Jamil, Madeleine Traverse, Alberto Tono, Bashir Sadjad, Lin Yang, Charles Lau, et~al.
\newblock {MedGemma} 1.5 technical report.
\newblock \emph{arXiv preprint arXiv:2604.05081}, 2026{\natexlab{a}}.

\bibitem[Sellergren et~al.(2026{\natexlab{b}})Sellergren, Kazemzadeh, Mahvar, Kiraly, Traverse, Kohlberger, Gao, Jaroensri, Xu, Jamil, et~al.]{sellergren2025medgemma}
Andrew Sellergren, Sahar Kazemzadeh, Fereshteh Mahvar, Atilla Kiraly, Madeleine Traverse, Timo Kohlberger, Chufan Gao, Ronnachai Jaroensri, Shawn Xu, Fayaz Jamil, et~al.
\newblock An open vision-language model for diverse medical applications.
\newblock \emph{Nature Medicine}, 32\penalty0 (10):\penalty0 3695--3704, 2026{\natexlab{b}}.

\bibitem[She et~al.(2026)She, Lu, Chen, Wang, and Huang]{she2026echovlm}
Chaoyin She, Ruifang Lu, Lida Chen, Wei Wang, and Qinghua Huang.
\newblock {EchoVLM}: Dynamic mixture-of-experts vision-language model for universal ultrasound intelligence.
\newblock In \emph{ACL}, 2026.

\bibitem[Sreenivas et~al.(2026)Sreenivas, Hanasoge, Yang, Taghibakhshi, Muralidharan, Aithal, and Molchanov]{sreenivas2026xtoken}
Sharath~Turuvekere Sreenivas, Adithyakrishna~Venkatesh Hanasoge, Mingyu Yang, Ali Taghibakhshi, Saurav Muralidharan, Ashwath Aithal, and Pavlo Molchanov.
\newblock {X-Token}: Projection-guided cross-tokenizer knowledge distillation.
\newblock \emph{arXiv preprint arXiv:2605.21699}, 2026.

\bibitem[Sun et~al.(2026)Sun, Zheng, Song, Zhong, Cheng, Feng, Liu, Fang, and Wang]{sun2026simct}
Jie Sun, Mao Zheng, Mingyang Song, Qiyong Zhong, Yilin Cheng, Bichuan Feng, Pengfei Liu, Junfeng Fang, and Xiang Wang.
\newblock {SimCT}: Recovering lost supervision for cross-tokenizer on-policy distillation.
\newblock In \emph{NeurIPS}, 2026.

\bibitem[Sun et~al.(2024)Sun, Wu, Zhu, Zheng, Chen, Zhang, Zhang, Wan, Lan, Zheng, Li, Lyu, Lin, and Yang]{sun2024pathmmu}
Yuxuan Sun, Hao Wu, Chenglu Zhu, Sunyi Zheng, Qizi Chen, Kai Zhang, Yunlong Zhang, Dan Wan, Xiaoxiao Lan, Mengyue Zheng, Jingxiong Li, Xinheng Lyu, Tao Lin, and Lin Yang.
\newblock {PathMMU}: A massive multimodal expert-level benchmark for understanding and reasoning in pathology.
\newblock In \emph{ECCV}, 2024.

\bibitem[Sun et~al.(2025)Sun, Si, Zhu, Gong, Zhang, Chen, Zhang, Shui, Lin, and Yang]{sun2025cpathomni}
Yuxuan Sun, Yixuan Si, Chenglu Zhu, Xuan Gong, Kai Zhang, Pingyi Chen, Ye Zhang, Zhongyi Shui, Tao Lin, and Lin Yang.
\newblock {CPath-Omni}: A unified multimodal foundation model for patch and whole slide image analysis in computational pathology.
\newblock In \emph{CVPR}, 2025.

\bibitem[Wan et~al.(2024)Wan, Huang, Cai, Quan, Bi, and Shi]{wan2024fusellm}
Fanqi Wan, Xinting Huang, Deng Cai, Xiaojun Quan, Wei Bi, and Shuming Shi.
\newblock Knowledge fusion of large language models.
\newblock In \emph{ICLR}, 2024.

\bibitem[Wang et~al.(2026)Wang, Yuan, Zhong, Zhang, Xiao, Sun, and Qi]{wang2026bpm}
Hao Wang, Kun Yuan, Wenlin Zhong, Minglei Zhang, Han Xiao, Ming Sun, and Honggang Qi.
\newblock Cross-tokenizer on-policy distillation via byte-prefix marginalization.
\newblock \emph{arXiv preprint arXiv:2607.22334}, 2026.

\bibitem[Wang et~al.(2025)Wang, Gao, Gu, Pu, Cui, Wei, Liu, Jing, Ye, Shao, et~al.]{wang2025internvl35}
Weiyun Wang, Zhangwei Gao, Lixin Gu, Hengjun Pu, Long Cui, Xingguang Wei, Zhaoyang Liu, Linglin Jing, Shenglong Ye, Jie Shao, et~al.
\newblock {InternVL3.5}: Advancing open-source multimodal models in versatility, reasoning, and efficiency.
\newblock \emph{arXiv preprint arXiv:2508.18265}, 2025.

\bibitem[Wei et~al.(2025)Wei, Qian, and Li]{wei2025funbench}
Qijie Wei, Kaiheng Qian, and Xirong Li.
\newblock {FunBench}: Benchmarking fundus reading skills of {MLLMs}.
\newblock In \emph{MICCAI}, 2025.

\bibitem[Wu et~al.(2026)Wu, Yao, Su, Ma, Zhang, Zhou, Mao, Geng, Tang, Duanmu, Chen, Chen, Fan, and Zhou]{wu2026ophthareason}
Ruiqi Wu, Yuang Yao, Na Su, Tengfei Ma, Chenran Zhang, Tao Zhou, Tianyu Mao, Ye Geng, Nianfeng Tang, Xiangyuan Duanmu, Qiqi Chen, Geng Chen, Wen Fan, and Yi Zhou.
\newblock Bridging the gap in ophthalmic {AI}: {MM-Retinal-Reason} dataset and {OphthaReason} model toward dynamic multimodal reasoning.
\newblock \emph{IEEE Transactions on Medical Imaging}, 2026.
\newblock Early access.

\bibitem[{Xiaomi LLM-Core Team} et~al.(2026){Xiaomi LLM-Core Team}, Xiao, Xia, Yang, Gao, Shen, Zhang, He, Lou, Luo, et~al.]{xiaomi2026mimo}
{Xiaomi LLM-Core Team}, Bangjun Xiao, Bingquan Xia, Bo Yang, Bofei Gao, Bowen Shen, Chen Zhang, Chenhong He, Chiheng Lou, Fuli Luo, et~al.
\newblock {MiMo-V2-Flash} technical report.
\newblock \emph{arXiv preprint arXiv:2601.02780}, 2026.

\bibitem[Xu et~al.(2026)Xu, Chan, Li, Aljunied, Yuan, Wang, Xiao, Chen, Liu, Li, Sun, Shen, Wang, Hu, Pan, Tan, Xu, Zhao, Zhang, and Rong]{xu2025lingshu}
Weiwen Xu, Hou~Pong Chan, Long Li, Mahani Aljunied, Ruifeng Yuan, Jianyu Wang, Chenghao Xiao, Guizhen Chen, Chaoqun Liu, Zhaodonghui Li, Yu Sun, Junao Shen, Chaojun Wang, Can Hu, Siwei Pan, Jie Tan, Tingyang Xu, Deli Zhao, Hao Zhang, and Yu Rong.
\newblock Lingshu: Generalist foundation model for unified multimodal medical understanding and reasoning.
\newblock \emph{IEEE TPAMI}, 2026.
\newblock Early access.

\bibitem[Yang et~al.(2025)Yang, Li, Yang, Zhang, Hui, Zheng, Yu, Gao, Huang, Lv, et~al.]{yang2025qwen3}
An Yang, Anfeng Li, Baosong Yang, Beichen Zhang, Binyuan Hui, Bo Zheng, Bowen Yu, Chang Gao, Chengen Huang, Chenxu Lv, et~al.
\newblock {Qwen3} technical report.
\newblock \emph{arXiv preprint arXiv:2505.09388}, 2025.

\bibitem[Zhang et~al.(2020)Zhang, Kishore, Wu, Weinberger, and Artzi]{zhang2020bertscore}
Tianyi Zhang, Varsha Kishore, Felix Wu, Kilian~Q. Weinberger, and Yoav Artzi.
\newblock {BERTScore}: Evaluating text generation with {BERT}.
\newblock In \emph{ICLR}, 2020.

\bibitem[Zhao et~al.(2026)Zhao, Hu, Zhu, Li, Liang, Shi, Lu, and Dou]{zhao2026opod}
Tong Zhao, Yuyang Hu, Yutao Zhu, Reed Li, Haijin Liang, Haibo Shi, Yu Lu, and Zhicheng Dou.
\newblock {OPOD}: On-policy omni distillation.
\newblock \emph{arXiv preprint arXiv:2607.20918}, 2026.

\end{thebibliography}
}

\clearpage
\SuppRefsStart

\twocolumn[\vbox{\hsize\textwidth\centering\vspace*{28pt}{\Large\textbf{Appendix}\par}\vspace{29.7872pt}}]
\appendix
\raggedbottom
\makeatletter\setlength{\@dblfptop}{0pt}\setlength{\@dblfpsep}{\dblfloatsep}\makeatother
\makeatletter\setlength{\@dblfpsep}{12pt plus 1fil}\setlength{\@dblfpbot}{0pt}\makeatother
\makeatletter\let\supp@ntf\NeedsTeXFormat\let\supp@pp\ProvidesPackage\def\NeedsTeXFormat#1[#2]{}\def\ProvidesPackage#1[#2]{}%
\input{stfloats.sty}\let\NeedsTeXFormat\supp@ntf\let\ProvidesPackage\supp@pp\makeatother
\setcounter{dbltopnumber}{4}\renewcommand{\dbltopfraction}{0.8}\renewcommand{\dblfloatpagefraction}{0.5}%
\setcounter{topnumber}{3}\setcounter{bottomnumber}{2}\renewcommand{\bottomfraction}{0.9}\makeatletter\setlength{\@fptop}{0pt}\setlength{\@fpsep}{\floatsep}\makeatother\renewcommand{\topfraction}{0.9}\renewcommand{\textfraction}{0.07}\renewcommand{\floatpagefraction}{0.8}%
\makeatletter\renewcommand\paragraph{\@startsection{paragraph}{4}{\z@}{1.5ex \@plus.5ex \@minus.2ex}{-1em}{\normalfont\normalsize\bfseries}}\makeatother

\section{HuluOPDCorpus and HuluOPDBench}
\label{sec:supp_datasets}

\paragraph{Sources.}
HuluOPDCorpus takes its prompts from public datasets of ultrasound, pathology, OCT, and fundus photography and, for dental imaging, from a subset of an in-house dental dataset; from it we use only the questions (\cref{tab:supp_src_corpus}).
HuluOPDBench takes its test items from held-out test splits of public datasets of the same four modalities (\cref{tab:supp_src_bench}).
From most of these datasets we take a subset of the images and questions rather than the full release, and the two tables give the number of prompts or test items taken from each.

\paragraph{Task groups.}
Every prompt and test item is assigned to one of four task groups (\cref{tab:supp_comp}).
Diagnosis and grading asks for a disease or its severity.
Recognition asks what the image shows, such as an organ, a tissue type, the imaging view, or a stain.
Detection and localization asks whether a finding is present, where it is, or how many there are.
Report and caption generation asks for a free-text description.
Within each group, \cref{fig:data} further divides the items by disease, organ, or sub-task.

\paragraph{Held-out verification.}
HuluOPDBench is strictly held out from HuluOPDCorpus, also where the two draw on the same public dataset.
We verified this image by image, comparing all 694k distinct prompt images with all 89k distinct test images of HuluOPDBench and the third-party benchmarks.
No image file occurs on both sides, and no HuluOPDBench image has a duplicate or near-duplicate among the prompts, so every HuluOPDBench score is measured on images outside the distillation data.
From the third-party benchmarks, we evaluate the items that pass the same check (\cref{tab:supp_bench}).

\begin{table}[!t]
  \centering
  \caption{\textbf{Composition} of HuluOPDCorpus and HuluOPDBench by modality and task group. DG: diagnosis and grading; REC: recognition; DET: detection and localization; GEN: report and caption generation.}
  \label{tab:supp_comp}
  \scriptsize
  \setlength{\tabcolsep}{0pt}
  \begin{tabular*}{\linewidth}{@{\extracolsep{\fill}}lrrrrr@{}}
  \toprule
  Modality & DG & REC & DET & GEN & Total \\
  \midrule
  \multicolumn{6}{@{}l}{\emph{HuluOPDCorpus (prompts)}} \\
  Ultrasound & 75,626 & 162,334 & 1,647 & 60,393 & 300,000 \\
  Pathology & 62,684 & 174,649 & 64,877 & 73,403 & 375,613 \\
  OCT & 74,727 & 11,662 & 212 & 207 & 86,808 \\
  Fundus photography & 71,800 & 18,355 & 11,425 & 7,308 & 108,888 \\
  Dental imaging & 31,478 & -- & 11,557 & -- & 43,035 \\
  Total & 316,315 & 367,000 & 89,718 & 141,311 & 914,344 \\
  \midrule
  \multicolumn{6}{@{}l}{\emph{HuluOPDBench (test items)}} \\
  Ultrasound & 5,444 & 1,795 & 112 & 2,276 & 9,627 \\
  Pathology & 5,881 & 14,744 & 2,952 & 999 & 24,576 \\
  OCT & 3,513 & 327 & 16 & 50 & 3,906 \\
  Fundus photography & 6,551 & 851 & 630 & 950 & 8,982 \\
  Total & 21,389 & 17,717 & 3,710 & 4,275 & 47,091 \\
  \bottomrule
  \end{tabular*}
\end{table}

\section{Data Selection}
\label{sec:supp_data}

\Cref{tab:supp_budget} lists the prompts used in each stage.

\paragraph{Single-specialist distillation.}
Each warm-up subset is balanced over task families and data sources; on-policy distillation uses every prompt of the modality.

\paragraph{Unified distillation.}
The unified warm-up mixes cached responses from the single-specialist warm-up data, balanced over tasks within each modality.
The unified prompt set for on-policy distillation holds all dental prompts and part of the prompts of the other four modalities; within a task, prompts are balanced over data sources, and we prefer prompts outside the unified warm-up.

\paragraph{Loss ablation.}
The loss ablation (\cref{sec:ablation}) uses 10k prompts per modality from HuluOPDCorpus, balanced over tasks and data sources, again preferring prompts outside \mbox{the unified warm-up}.

\section{Implementation Details}
\label{sec:supp_impl}

\paragraph{Models and training settings.}
\Cref{tab:models} lists the teachers and students, and \cref{tab:supp_hparams} the settings of the two training stages.
We fine-tuned the ultrasound, pathology, OCT, and fundus photography teachers with all parameters for three epochs on 29k--380k curated image--question--answer examples per modality.
In the student's warm-up, truncated, empty, or strongly repetitive teacher responses are dropped before fine-tuning.

\paragraph{On-policy distillation.}
The student samples its responses with a colocated vLLM~\cite{kwon2023vllm} engine, and its weights are copied to the engine after every optimizer step.
Unlike the warm-up, on-policy distillation updates all student parameters, including the vision encoder, so the teacher's supervision also adapts the student's visual features.
In the unified run, every step contains prompts of all five modalities in their shares of the unified prompts (\cref{tab:supp_budget}), and the two epochs take 10,760 steps.
Every model is trained once.
All experiments run on \mbox{NVIDIA A800 (80GB) GPUs}.

\paragraph{Numerical stability.}
We compute the loss in the log domain, taking $\log p_S$ and $\log p_T$ from a log-softmax in FP32 and applying no floor or clamp to the merged probabilities; the two models themselves run in BF16.

\section{Evaluation Details}
\label{sec:supp_eval}
\setlength{\dbltextfloatsep}{8pt plus 2pt minus 2pt}%

\begin{table*}[!b]
  \centering
  \caption{\textbf{Teachers and students.} $^\dagger$Fine-tuned by us. Params: total parameters, with the vision encoder's share in parentheses. Vocab.: tokenizer entries, including special tokens.}
  \label{tab:models}
  \scriptsize
  \setlength{\tabcolsep}{0pt}
  \begin{tabular*}{\textwidth}{@{\extracolsep{\fill}}lllllr@{}}
  \toprule
  Model & Teacher for & Vision encoder & Language model & Vocab. & Params \\
  \midrule
  Qwen3.5\mbox{-}9B$^\dagger$ & Ultrasound, pathology & Qwen3.5 ViT (0.46B) & Qwen3.5\mbox{-}9B & 248,077 & 9.41B \\
  InternVL3.5\mbox{-}8B$^\dagger$ & OCT, fundus photography & InternViT (0.30B) & Qwen3-8B & 151,679 & 8.53B \\
  DentVLM~\cite{meng2025dentvlm} & Dental imaging & Qwen2-VL ViT (0.68B) & Qwen2-7B & 151,657 & 8.29B \\
  \midrule
  HuluMed\mbox{-}4B~\cite{jiang2025hulumed} & -- (student) & Qwen3-VL ViT (0.42B) & Qwen3-4B & 151,669 & 4.44B \\
  Qwen3.5\mbox{-}0.8B~\cite{qwen35} & -- (student) & Qwen3.5 ViT (0.10B) & Qwen3.5\mbox{-}0.8B & 248,070 & 0.85B \\
  \bottomrule
  \end{tabular*}
  \vspace{2pt}
  \caption{\textbf{Source datasets of HuluOPDCorpus}, with the number of prompts taken from each.}
  \label{tab:supp_src_corpus}
  \scriptsize
  \setlength{\tabcolsep}{3pt}
  \begin{tabular*}{\textwidth}{@{\extracolsep{\fill}}l r p{0.80\textwidth}@{}}
  \toprule
  Modality & Prompts & Source datasets (prompts taken from each) \\
  \midrule
  Ultrasound & 300,000 & SonoInstruct~\cite{wang2026sonoinstruct} (117,516), RadImageNet~\cite{mei2022radimagenet} (54,389), PMC-OA~\cite{lin2023pmcoa} (36,570), EchoVLM-Dataset~\cite{she2026echovlm} (13,746), ROCOv2~\cite{ruckert2024rocov2} (10,724), PMC-VQA~\cite{zhang2024pmcvqa} (9,023), Knee Classification~\cite{nelson2023knee} (7,803), ThyroidUS Pathology~\cite{hou2024thyroidus} (5,719), UIdataGB~\cite{turki2024uidatagb} (4,471), SMC-LUD~\cite{tak2026smclud} (3,722), TN5000~\cite{zhang2025tn5000} (3,678), FETAL\_PLANES\_DB~\cite{burgosartizzu2020fetalplanes} (3,175), UBIBC~\cite{sairam2022ubibc} (3,166), GDPH\&SYSUCC~\cite{mo2023gdphsysucc} (3,149), SegThy~\cite{kronke2022segthy} (3,089), TN3K~\cite{gong2023tn3k} (2,476), BEHSOF (1,436), BUS-BRA~\cite{gomezflores2024busbra} (1,414), Knee Grading Ultrasound Classification Dataset~\cite{nelson2023knee} (1,394), CVA-Net~\cite{lin2022cvanet} (1,330), LEPset~\cite{li2024lepset} (1,313), PCOS~\cite{choudhari2022pcos} (1,295), RegensburgAppendicitis~\cite{marcinkevics2024appendicitis} (1,291), MMOTU~\cite{zhao2022mmotu} (1,099), UUSIVC2026 (803), COVID-BLUeS~\cite{wiedemann2025covidblues} (699), CAMUS~\cite{leclerc2019camus} (694), BUS-CoT~\cite{yu2026buscot} (557), PCOSGen v2~\cite{handa2025pcosgen} (515), US-Liver~\cite{lu2025usliver} (511), BUS-UC~\cite{iqbal2024busuc} (463), POCUS~\cite{born2021pocus} (439), FPUS23~\cite{prabakaran2023fpus23} (416), Lung-Ultrasound-AI-Disease-Classification~\cite{katumba2025lungus} (379), JNU-IFM~\cite{lu2022jnuifm} (371), AUITD~\cite{maroua2022auitd} (287), BUSI~\cite{aldhabyani2020busi} (278), MUPSD~\cite{jiang2024mupsd} (225), AUL~\cite{xu2023aul} (142), Dermatologic Ultrasound~\cite{laverdesaad2022dermus} (137), KFGNet~\cite{wang2022kfgnet} (66), Carotid Ultrasound~\cite{choudhary2024carotid} (25), BrEaST~\cite{pawlowska2024breast} (3), BUS-UCLM~\cite{vallez2025busuclm} (2) \\
  \midrule
  Pathology & 375,613 & SkinCancer~\cite{kriegsmann2022skincancer} (64,227), CRC-100K~\cite{kather2019crc100k} (58,286), Quilt-Instruct~\cite{seyfioglu2024quiltllava} (45,165), Quilt-1M~\cite{ikezogwo2023quilt1m} (45,112), PCam~\cite{veeling2018pcam} (45,066), LC25000~\cite{borkowski2019lc25000} (30,818), ARCH-Open~\cite{naeem2025pathrag} (29,307), PMC-VQA~\cite{zhang2024pmcvqa} (28,787), DigestPath 2019~\cite{da2022digestpath} (14,223), ARCH~\cite{gamper2021arch} (5,944), PathVQA~\cite{he2020pathvqa} (4,868), WSSS4LUAD~\cite{han2022wsss4luad} (1,965), SICAPv2~\cite{silvarodriguez2020sicapv2} (1,412), MHIST~\cite{wei2021mhist} (241), Osteosarcoma~\cite{arunachalam2019osteosarcoma} (192) \\
  \midrule
  OCT & 86,808 & OCT2017~\cite{kermany2018oct2017} (80,724), OCTDL~\cite{kulyabin2024octdl} (2,169), RETOUCH~\cite{bogunovic2019retouch} (1,305), PMC-VQA~\cite{zhang2024pmcvqa} (1,123), DeepEyeNet~\cite{huang2021deepeyenet} (577), MM-Retinal~\cite{wu2024mmretinal} (404), OCTID~\cite{gholami2020octid} (340), MultiEYE~\cite{wang2025multieye} (97), NEH~\cite{sotoudehpaima2022neh} (69) \\
  \midrule
  Fundus photography & 108,888 & MultiEYE~\cite{wang2025multieye} (51,947), DeepEyeNet~\cite{huang2021deepeyenet} (18,912), ODIR-5K~\cite{li2021odir} (9,063), HPMI~\cite{huang2024hpmi} (6,298), DME-VQA~\cite{tasconmorales2022dmevqa} (3,817), MM-Retinal~\cite{wu2024mmretinal} (3,032), Glaucoma Fundus~\cite{ahn2018glaucoma} (1,941), APTOS~\cite{karthik2019aptos} (1,789), DDR~\cite{li2019ddr} (1,640), RFMiD~\cite{pachade2021rfmid} (1,235), IDRiD~\cite{porwal2018idrid} (1,221), Retinal-Lesions~\cite{wei2021retinallesions} (1,087), e-ophtha~\cite{decenciere2013eophtha} (1,044), PMC-VQA~\cite{zhang2024pmcvqa} (1,009), ACRIMA~\cite{diazpinto2019acrima} (904), JSIEC~\cite{cen2021jsiec} (839), RIM-ONE DL~\cite{fumerobatista2020rimonedl} (836), PALM~\cite{fang2024palm} (701), DeepDRiD~\cite{liu2022deepdrid} (436), PAPILA~\cite{kovalyk2022papila} (430), DIARETDB0~\cite{kauppi2006diaretdb0} (260), Messidor~\cite{decenciere2014messidor} (211), Retina Diagnosis~\cite{chen2017retinadataset} (132), Messidor-2~\cite{decenciere2014messidor,abramoff2013messidor2} (66), Retino~\cite{wang2023medfmc} (38) \\
  \midrule
  Dental imaging & 43,035 & In-house dental data (43,035) \\
  \bottomrule
  \end{tabular*}
  \vspace{2pt}
  \caption{\textbf{Source datasets of HuluOPDBench}, with the number of test items taken from each (for dental imaging, see \cref{tab:supp_bench}).}
  \label{tab:supp_src_bench}
  \scriptsize
  \setlength{\tabcolsep}{3pt}
  \begin{tabular*}{\textwidth}{@{\extracolsep{\fill}}l r p{0.80\textwidth}@{}}
  \toprule
  Modality & Items & Source datasets (test items taken from each) \\
  \midrule
  Ultrasound & 9,627 & EchoVLM-Dataset~\cite{she2026echovlm} (1,544), SonoInstruct~\cite{wang2026sonoinstruct} (1,336), UIdataGB~\cite{turki2024uidatagb} (1,206), ThyroidUS Pathology~\cite{hou2024thyroidus} (812), PMC-VQA~\cite{zhang2024pmcvqa} (727), SMC-LUD~\cite{tak2026smclud} (720), PMC-OA~\cite{lin2023pmcoa} (578), ROCOv2~\cite{ruckert2024rocov2} (569), TN5000~\cite{zhang2025tn5000} (384), TN3K~\cite{gong2023tn3k} (326), BUS-BRA~\cite{gomezflores2024busbra} (242), Lung-Ultrasound-AI-Disease-Classification~\cite{katumba2025lungus} (204), UUSIVC2026 (202), RegensburgAppendicitis~\cite{marcinkevics2024appendicitis} (152), BUS-UC~\cite{iqbal2024busuc} (144), AUITD~\cite{maroua2022auitd} (102), US-Liver~\cite{lu2025usliver} (90), MMOTU~\cite{zhao2022mmotu} (88), BUSI~\cite{aldhabyani2020busi} (81), AUL~\cite{xu2023aul} (60), BEHSOF (60) \\
  \midrule
  Pathology & 24,576 & ESCA~\cite{tolkach2023esca} (5,939), PCam~\cite{veeling2018pcam} (2,899), CRC-100K~\cite{kather2019crc100k} (2,520), BreakHis~\cite{spanhol2016breakhis} (1,598), SkinCancer~\cite{kriegsmann2022skincancer} (1,570), MHIST~\cite{wei2021mhist} (1,561), RenalCell~\cite{brummer2023renalcell} (1,500), WSSS4LUAD~\cite{han2022wsss4luad} (1,499), Chaoyang~\cite{zhu2022chaoyang} (1,394), Osteosarcoma~\cite{arunachalam2019osteosarcoma} (992), Databiox~\cite{bolhasani2020databiox} (922), Quilt-1M~\cite{ikezogwo2023quilt1m} (847), PathMMU~\cite{sun2024pathmmu} (380), PMC-VQA~\cite{zhang2024pmcvqa} (327), BACH~\cite{aresta2019bach} (269), Quilt-VQA~\cite{seyfioglu2024quiltllava} (180), ARCH~\cite{gamper2021arch} (150), ARCH-Open~\cite{naeem2025pathrag} (29) \\
  \midrule
  OCT & 3,906 & OCT2017~\cite{kermany2018oct2017} (2,191), OCTDL~\cite{kulyabin2024octdl} (940), OCTID~\cite{gholami2020octid} (343), MultiEYE~\cite{wang2025multieye} (265), DeepEyeNet~\cite{huang2021deepeyenet} (101), PMC-VQA~\cite{zhang2024pmcvqa} (66) \\
  \midrule
  Fundus photography & 8,982 & ODIR-5K~\cite{li2021odir} (1,576), MultiEYE~\cite{wang2025multieye} (1,405), Retino~\cite{wang2023medfmc} (1,201), APTOS~\cite{karthik2019aptos} (1,180), DeepEyeNet~\cite{huang2021deepeyenet} (663), MM-Retinal~\cite{wu2024mmretinal} (523), Messidor~\cite{decenciere2014messidor} (362), HPMI~\cite{huang2024hpmi} (340), JSIEC~\cite{cen2021jsiec} (339), Glaucoma Fundus~\cite{ahn2018glaucoma} (316), Retina Diagnosis~\cite{chen2017retinadataset} (220), DeepDRiD~\cite{liu2022deepdrid} (209), ACRIMA~\cite{diazpinto2019acrima} (200), IDRiD~\cite{porwal2018idrid} (118), RIM-ONE DL~\cite{fumerobatista2020rimonedl} (103), PAPILA~\cite{kovalyk2022papila} (101), PMC-VQA~\cite{zhang2024pmcvqa} (67), e-ophtha~\cite{decenciere2013eophtha} (59) \\
  \bottomrule
  \end{tabular*}
\end{table*}

\paragraph{Scoring and decoding.}
The judge reads the question, the reference answer, and the full response, without the image, and scores only the final answer.
BERTScore uses layer 14 of Longformer\mbox{-}large~\cite{beltagy2020longformer}, which reads long reports without truncation.
All models answer with greedy decoding, a repetition penalty of 1.05, and at most 2,048 new tokens.

\section{Loss Ablation Details}
\label{sec:supp_ablation}

Each variant of \cref{tab:ablation} trains for two epochs with the settings of \cref{sec:exp_five_to_one}, computes its loss in FP32, and is evaluated on all benchmarks.
On OCT, fundus photography, and dental imaging, C scores $-$0.25, $-$0.06, and $+$0.18 points relative to B, and D $-$0.13, $+$0.01, and $+$0.16.

With positional pairing, 58\% of the student's tokens on pathology receive a teacher prediction for a different part of the text (measured on warm-up responses).
Variant~B then degenerates into repetition: 11.7\% of its pathology responses run into the 2,048-token limit, against 0.1\% for the other three variants, which lowers its pathology score in \cref{tab:ablation}.
\Cref{fig:ablation_examples} shows the four variants on one pathology question that their warm-up checkpoint answers correctly: C and D give the same short, correct response, A rules out smooth muscle and answers incorrectly, and B repeats word fragments until the token limit.

\begin{table*}[p]
  \centering
  \caption{\textbf{Prompts used in each stage}, by modality. OPD: on-policy distillation; single-specialist OPD uses all HuluOPDCorpus prompts of the modality. Ablation: loss ablation of \cref{sec:ablation}.}
  \label{tab:supp_budget}
  \scriptsize
  \setlength{\tabcolsep}{0pt}
  \begin{tabular*}{\linewidth}{@{\extracolsep{\fill}}lrrrrr@{}}
  \toprule
   & \multicolumn{2}{c}{Single-specialist} & \multicolumn{2}{c}{Unified} & \\
  \cmidrule(l{6pt}){2-3}\cmidrule(l{6pt}){4-5}
  Modality & Warm-up & OPD & Warm-up & OPD & Ablation \\
  \midrule
  Ultrasound & 50k & 300,000 & 23k & 98,980 & 10k \\
  Pathology & 50k & 375,613 & 23k & 98,980 & 10k \\
  OCT & 20k & 86,808 & 10k & 43,035 & 10k \\
  Fundus photography & 20k & 108,888 & 14k & 60,249 & 10k \\
  Dental imaging & 10k & 43,035 & 10k & 43,035 & 10k \\
  \midrule
  Total & 150k & 914,344 & 80k & 344,279 & 50k \\
  \bottomrule
  \end{tabular*}
\end{table*}

\begin{table*}[p]
  \centering
  \caption{\textbf{Training settings} of the two stages of HuluOPD. Exceptions are given in parentheses.}
  \label{tab:supp_hparams}
  \scriptsize
  \setlength{\tabcolsep}{3pt}
  \renewcommand{\arraystretch}{1.05}
  \begin{tabular*}{\linewidth}{@{}p{0.12\linewidth}@{\extracolsep{\fill}}p{0.48\linewidth}p{0.36\linewidth}@{}}
  \toprule
  \raggedright Setting & \raggedright Warm-up & \raggedright On-policy distillation \tabularnewline
  \midrule
  \raggedright Responses & \raggedright from the teacher; greedy decoding, repetition penalty 1.05 & \raggedright sampled by the student; temperature 0.7, top-$p$ 0.8, top-$k$ 20, repetition penalty 1.08 \tabularnewline[2pt]
  \raggedright Response length & \raggedright at most 2,048 new tokens (768 for DentVLM) & \raggedright at most 512 new tokens (1,024 in the unified run) \tabularnewline[2pt]
  \raggedright Trained parameters & \raggedright language model and projector; vision encoder frozen & \raggedright all, including the vision encoder \tabularnewline[2pt]
  \raggedright Epochs & \raggedright 1 & \raggedright 2 \tabularnewline[2pt]
  \raggedright Global batch & \raggedright 32 (16 for dental imaging) & \raggedright 64 \tabularnewline[2pt]
  \raggedright Learning rate & \raggedright $10^{-5}$ for HuluMed\mbox{-}4B; $2\times10^{-5}$ for Qwen3.5\mbox{-}0.8B ($10^{-5}$ for its projector) & \raggedright $2\times10^{-7}$ \tabularnewline[2pt]
  \raggedright Schedule & \raggedright cosine with 3\% linear warm-up & \raggedright constant after 3\% warm-up \tabularnewline
  \bottomrule
  \end{tabular*}
\end{table*}

\begin{table*}[p]
  \centering
  \caption{\textbf{Unified students by task group}, over all benchmarks. Task groups as in \cref{tab:supp_comp}; DG, REC, and DET are scored as the Avg.\ of \cref{tab:unified}, GEN by BERTScore; green: gain over Base.}
  \label{tab:taskgroup}
  \scriptsize
  \setlength{\tabcolsep}{0pt}
  \begin{tabular*}{\linewidth}{@{\extracolsep{\fill}}lrccccc@{}}
  \toprule
   & & & \multicolumn{2}{c}{HuluMed\mbox{-}4B} & \multicolumn{2}{c}{Qwen3.5\mbox{-}0.8B} \\
  \cmidrule(l{4pt}){4-5}\cmidrule(l{4pt}){6-7}
  Group & Items & Teacher & Base & + HuluOPD & Base & + HuluOPD \\
  \midrule
  DG & 53,958 & 69.28 & 48.70 & \g{65.40}{+16.70} & 40.39 & \g{61.92}{+21.53} \\
  REC & 52,840 & 84.61 & 66.24 & \g{77.59}{+11.35} & 62.60 & \g{72.37}{+9.77} \\
  DET & 19,994 & 82.32 & 62.96 & \g{80.65}{+17.69} & 65.36 & \g{78.87}{+13.51} \\
  GEN & 5,073 & 87.76 & 83.16 & \g{86.53}{+3.37} & 81.63 & \g{86.12}{+4.49} \\
  \bottomrule
  \end{tabular*}
\end{table*}

\begin{table*}[p]
  \centering
  \caption{\textbf{Third-party benchmarks}: the parts we evaluate and the parts we leave out. Items: test items evaluated, after the check of \cref{sec:supp_datasets}. Task names follow each benchmark.}
  \label{tab:supp_bench}
  \scriptsize
  \setlength{\tabcolsep}{4pt}
  \renewcommand{\arraystretch}{1.05}
  \begin{tabular*}{\textwidth}{@{\extracolsep{\fill}}l l r p{0.33\textwidth} p{0.29\textwidth}@{}}
  \toprule
  Modality & Benchmark & Items & Evaluated & Left out \\
  \midrule
  Ultrasound & OmniMedVQA\mbox{-}US~\cite{hu2024omnimedvqa} & 10,976 & ultrasound questions & other imaging modalities \\
   & U2\mbox{-}BENCH~\cite{le2026u2bench} & 3,513 & disease diagnosis; view recognition and assessment; report generation; caption generation & lesion localization; organ detection; keypoint detection; clinical value estimation; plane-quality questions on ACOUSLIC-AI; fetal-orientation questions on FPUS23 \\
  \midrule
  Pathology & PathMMU~\cite{sun2024pathmmu} & 8,202 & multiple-choice questions whose images are publicly available & questions whose images are not available \\
   & PathVQA~\cite{he2020pathvqa} & 2,681 & yes/no (1,552) and open-ended (1,129) questions on the 404 microscopy images of the test split & gross-pathology photographs; training and validation splits \\
  \midrule
  OCT & FunBench~\cite{wei2025funbench} & 15,773 & fine-grained modality perception (L1b); disease diagnosis (L4) & coarse modality perception (L1a); lesion analysis (L3); MMC-AMD \\
  \midrule
  Fundus photography & FunBench~\cite{wei2025funbench} & 23,154 & fine-grained modality perception (L1b); anatomy perception (L2); disease diagnosis (L4) & coarse modality perception (L1a); lesion analysis (L3); ultra-widefield images (TOP and part of DeepDRiD); MMC-AMD \\
  \midrule
  Dental imaging & DentVLM\mbox{-}Bench & 17,427 & yes/no finding and open-ended diagnosis questions on radiographs and intraoral photographs & -- \\
   & DentVLM\mbox{-}Clinical\mbox{-}Bench & 3,105 & as DentVLM\mbox{-}Bench & -- \\
  \bottomrule
  \end{tabular*}
\end{table*}

\providecommand{\cmark}{\textcolor{ForestGreen}{$\checkmark$}}
\providecommand{\xmark}{\textcolor{BrickRed}{$\times$}}
\begin{figure*}[t]
  \scriptsize
  \newcommand{\fnd}[1]{{\setlength{\fboxsep}{0pt}\colorbox{Goldenrod!80}{\vphantom{hg}\hspace{0.6pt}\textbf{#1}\hspace{0.6pt}}}}
  \newcommand{\den}[1]{{\setlength{\fboxsep}{0pt}\colorbox{BrickRed!42}{\vphantom{hg}\hspace{0.6pt}\textbf{#1}\hspace{0.6pt}}}}
  \newcommand{\rep}[1]{{\setlength{\fboxsep}{0pt}\colorbox{Orange!45}{\vphantom{hg}\hspace{0.6pt}\textbf{#1}\hspace{0.6pt}}}}
  \newcommand{\caserow}[2]{\noindent{\setlength{\fboxsep}{2.5pt}\colorbox{#1}{\parbox{\dimexpr\linewidth-5pt\relax}{\raggedright #2}}}\par\vspace{2pt}}
  \noindent\begin{minipage}[t]{0.11\linewidth}\vspace{0pt}\includegraphics[width=\linewidth]{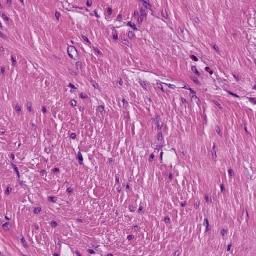}\end{minipage}\hfill
  \begin{minipage}[t]{0.87\linewidth}\vspace{0pt}\raggedright
    \textbf{Question.} What is the diagnosis/type of this histology image?
    A.~Adventitia; B.~Lamina Propria; C.~Muscularis Mucosae; D.~Muscularis Propria; E.~Regression Tumor; F.~Submucosa (Specific); G.~Squamous Epithelium; H.~Submucosa; I.~Submucosal Glands; J.~Tumor; K.~Ulcus.
    Explain your reasoning and then give the final answer.\par\vspace{3pt}
    \textbf{Reference.} D.~Muscularis Propria
  \end{minipage}\par\vspace{5pt}
  \caserow{black!7}{\textbf{Warm-up checkpoint}, start of all variants \cmark\ \emph{D.~Muscularis Propria}, 60 tokens: ``The image shows spindle-shaped cells with elongated nuclei arranged in fascicles, which is characteristic of \fnd{smooth muscle tissue}. Comparing this morphology to the provided options, the features align best with the muscularis propria layer of the gastrointestinal tract. Therefore, the diagnosis is Muscularis Propria.''}
  \caserow{BrickRed!11}{\textbf{A}, Position\,/\,Sorted \xmark\ \emph{F.~Submucosa (Specific)}, 768 tokens: ``1. Analyze the image features: [\ldots] There are no clear muscle fascicles with striations (\den{ruling out} smooth muscle layers like Muscularis Propria, though it shares some eosinophilia). [\ldots] D. Muscularis Propria: [\ldots] The image shows fibrous connective tissue, \den{not muscle bundles}. [\ldots] F. Submucosa (Specific): [\ldots] The image perfectly matches the histological appearance of the submucosa [\ldots] Therefore, the diagnosis is Submucosa (Specific).''}
  \caserow{Orange!13}{\textbf{B}, Position\,/\,Hybrid\,/\,Fixed \xmark\ no final answer, stopped at the 2,048-token limit: ``1. Analyze the image features: [\ldots] D. Muscularis Prop: [\ldots] The image shows bundles of eosinophilic fibers with elongated nuclei, \fnd{consistent with smooth muscle}. However, the texture is very fibrous and wavy [\ldots] F. Subm Submucosa (Specific): [\ldots] G. \rep{Squamous} \rep{Squamous} \rep{Squamous} \rep{Squamous} \rep{Squ} \rep{Squ} \rep{Squ} [\ldots, \emph{Squ} repeated until the limit]''}
  \caserow{RoyalBlue!11}{\textbf{C}, Text\,/\,Hybrid\,/\,Fixed \cmark\ \emph{D.~Muscularis Propria}, 63 tokens: ``The image shows spindle-shaped cells with eosinophilic cytoplasm arranged in fascicles, which is characteristic of \fnd{smooth muscle tissue}. Comparing this to the provided options, the features align best with the muscularis propria layer of the gastrointestinal tract. Therefore, the diagnosis is Muscularis Propria.''}
  \caserow{ForestGreen!14}{\textbf{D (ours)}, Text\,/\,Hybrid\,/\,Continuation \cmark\ \emph{D.~Muscularis Propria}, 63 tokens, the same response as C: ``The image shows spindle-shaped cells with eosinophilic cytoplasm arranged in fascicles, which is characteristic of \fnd{smooth muscle tissue}. Comparing this to the provided options, the features align best with the muscularis propria layer of the gastrointestinal tract. Therefore, the diagnosis is Muscularis Propria.''}
  \caption{\textbf{Responses of the loss variants} of \cref{tab:ablation} and of their common warm-up checkpoint to one pathology question from HuluOPDBench. {\setlength{\fboxsep}{1pt}\colorbox{Goldenrod!80}{Yellow}} marks where the student has learned the teacher's key finding, {\setlength{\fboxsep}{1pt}\colorbox{BrickRed!42}{dark red}} where it has not, and {\setlength{\fboxsep}{1pt}\colorbox{Orange!45}{orange}} repetition; [\ldots]: omitted text.}
  \label{fig:ablation_examples}
\end{figure*}

\makeatletter\par\newpage\if@firstcolumn\else\null\newpage\fi\makeatother
{
    \small
    \SuppReferences{ieeenat_fullname}{main}
}

\end{document}